\documentclass{article}

\usepackage[margin=1in]{geometry} 
\usepackage{graphicx}
\usepackage[utf8]{inputenc} %
\usepackage[T1]{fontenc}    %
\usepackage[pdfencoding=auto, psdextra,hidelinks]{hyperref}       %
\usepackage{url}            %
\usepackage{booktabs}       %
\usepackage{amsfonts}       %
\usepackage{nicefrac}       %
\usepackage{microtype}      %
\usepackage{xcolor}         %
\usepackage{amsmath}
\usepackage{amssymb}
\usepackage{amsthm}
\usepackage{mathtools}
\usepackage{bbm}
\usepackage{algorithm}
\usepackage{algorithmic}
\usepackage{subcaption}
\usepackage[textsize=tiny]{todonotes}
\usepackage[compress,sort]{natbib} 

\usepackage{thm-restate}

\theoremstyle{plain}
\newtheorem{theorem}{Theorem}[section]
\newtheorem{proposition}[theorem]{Proposition}

\newtheorem{corollary}[theorem]{Corollary}
\theoremstyle{definition}

\theoremstyle{remark}
\newtheorem{remark}[theorem]{Remark}

\usepackage[capitalize,noabbrev]{cleveref}
\usepackage{amartya_ltx}
\def\EE{{\mathbb E}}    %
\def\cD{{\mathcal D}}
\def\cQ{{\mathcal Q}}
\def\cR{{\mathcal R}}
\def\cX{{\mathcal X}}
\def\cY{{\mathcal Y}}
\def\cA{{\mathcal A}}
\def\cP{{\mathcal P}}
\def\cF{{\mathcal F}}

\usepackage{bbm}

\usepackage[textsize=tiny]{todonotes}

\newcommand{\hellotitle}{The Role of Learning Algorithms in Collective Action}

\title{\hellotitle{}}

\usepackage{authblk}
\stepcounter{footnote}
\addtocounter{footnote}{-1}
\author[1,2]{Omri Ben-Dov*}
\author[3]{Jake Fawkes*}
\author[1,2]{Samira Samadi}
\author[1]{Amartya Sanyal}
\affil[1]{Max Planck Institute for Intelligent Systems, T\"ubingen, Germany}
\affil[2]{T\"ubingen AI Center}
\affil[3]{Department of Statistics, University of Oxford}
\date{}        
\newcommand\equalContrib{%
  \begin{NoHyper}
  \def\thefootnote{*}\footnotetext{Equal contribution}%
  \addtocounter{footnote}{0}%
  \end{NoHyper}
  \def\thefootnote{\arabic{footnote}}
}

\newcommand\correspondence{%
  \begin{NoHyper}
  \def\thefootnote{}\footnotetext{Email: odov@tue.mpg.de,~jake.fawkes@stats.ox.ac.uk,~ssamadi@tuebingen.mpg.de,~amsa@di.ku.dk}%
  \addtocounter{footnote}{0}%
  \end{NoHyper}
  \def\thefootnote{\arabic{footnote}}
}

\newcommand\acceptance{%
  \begin{NoHyper}
  \def\thefootnote{$\dagger$}\footnotetext{Accepted at the International Conference in Machine Learning (ICML), 2024}%
  \addtocounter{footnote}{0}%
  \end{NoHyper}
  \def\thefootnote{\arabic{footnote}}
}

\begin{document}

\maketitle

\acceptance{}
\correspondence{}
\equalContrib{}

\begin{abstract}
\textit{Collective action} in machine learning is the study of the control that a coordinated group can have over machine learning algorithms. While previous research has concentrated on assessing the impact of collectives against Bayes (sub-)optimal classifiers, this perspective is limited in that it does not account for the choice of learning algorithm. Since classifiers seldom behave like Bayes classifiers and are influenced by the choice of learning algorithms along with their inherent biases, in this work we initiate the study of how the choice of the learning algorithm plays a role in the success of a collective in practical settings. Specifically, we focus on distributionally robust optimization (DRO), popular for improving a worst group error, and on the ubiquitous stochastic gradient descent (SGD), due to its inductive bias for ``simpler'' functions. Our empirical results, supported by a theoretical foundation, show that the effective size and success of the collective are highly dependent on properties of the learning algorithm. This highlights the necessity of taking the learning algorithm into account when studying the impact of collective action in machine learning.
\end{abstract}

\section{Introduction}
\label{sec:intro}

With the rapid increase in deployed machine learning models, a large number of firms rely on data contributed by users~\citep{gerlitz_like_2013} to train their algorithms. 
In response to this, consumers and users have searched for ways to alter their data to influence the outputs of such models~\citep{chen_thrown_2018,burrell_when_2019,rahman_invisible_2021}.
\textit{Algorithmic collective action} \citep{olson_logic_1965,hardt_algorithmic_2023} has emerged as a formal framework to study the effect a coordinated group of individuals can have on such models, by altering the data they provide to the firm.
This naturally leads to various technical questions regarding how large the collective needs to be, how effective different data altering strategies are, and which details about firm's algorithm can be leveraged by the collective to achieve the best results. 

\citet{hardt_algorithmic_2023} initiated the formal study into the success of different collective action strategies. Their work provided theoretical analysis based on the fractional size of the collective and on the properties of the signal with regards to the original data distribution. However, their results  only hold for Bayes (sub-)optimal classifiers which do not immediately adapt to peculiarities of practical learning algorithms. In this work, we investigate two types of commonly used learning algorithms that exhibit distinct properties: (1) Distributional robustness and (2) Simplicity bias. We show how these properties lead to significantly different levels of collective success that are unexplained by prior work.

When the data is composed of multiple sub-populations, algorithms that optimise for average performance often perform poorly in minority sub-populations~\citep{meinshausen_maximin_2015}.
A fairness-focused firm will want to ensure that their trained learning model performs well uniformly on all sub-populations, as opposed to being on-average good. Distributionally Robust Optimisation~(DRO) is a family of algorithms designed to maximize this per-group accuracy~\citep{hashimoto_fairness_2018,wang_robust_2020}.
We show that as a consequence, a small collective achieves higher success when the training algorithm performs DRO, compared to standard empirical risk minimization (ERM). Conversely, a large collective in the same settings achieves lower success, contradicting the expectation set by previous work regarding the effectiveness of large collectives.

Second, most machine learning algorithms used today are based on some form of gradient descent (GD). Such algorithms, including the popular Stochastic Gradient Descent~(SGD), Adam~\citep{kingma_adam_2015}, and RMSProp, exhibit a preference for learning functions that are ``simpler".
This inductive bias of GD algorithms is popularly referred to as simplicity bias~\citep{kalimeris_sgd_2019,shah_pitfalls_2020}.
In practice, this preference results in the model ``overlooking'' certain complex features.
We demonstrate that these overlooked features can be leveraged by a collective to design a strategy that will gain higher success compared to what is possible on a Bayes optimal classifier.

Our work initiates a study into an algorithm-dependent view on the success of collective action. We provide a theoretical foundation and empirical evidence to analyse the success of the collective for two important categories of algorithms: DRO and algorithms with simplicity bias~(e.g. SGD).

\section{Problem Formulation and Notation}

In this section, we provide a formal discussion of both Collective Action and Distributionally Robust Optimisation in addition to defining the various notations that will be used throughout this manuscript.

\subsection{Collective Action}\label{sec:coll-action}

While there are several goals in the domain of algorithmic collective action defined in~\citet{hardt_algorithmic_2023}, in this work we will focus on the goal of collectively \textit{planting a signal} with a \textit{features-label strategy}.
We believe that other important goals, like~\textit{erasing signals}, can also benefit from the techniques and observations of this paper, but we leave the detailed study of such settings to future work. 

\paragraph{Planting a signal} Given a base distribution \basedist{} on the domain of features and labels $\cX\times\cY$, the classifier observes the mixture distribution \begin{equation}\label{eq:coll-dist}
    \mixdist=\strength\colldist + \left(1-\strength\right)\basedist,
\end{equation}
where \strength{} is the collective's proportional size and \colldist{} is the collective's distribution. The goal of the collective is to create an association in a classifier $f:\cX\rightarrow\cY$, between a signal $\collfn:\cX\rightarrow\cX$ and a label $y^*\in\cY$. Formally, the collective's goal is to maximize the success defined as
\begin{equation}S\left(\strength\right)=\basedist\left[f\left(\collfn\left(x\right)\right)=y^{*}\right].
\end{equation}

The collective modifies their own data by planting a signal $\left(x,y\right){\rightarrow}\left(g\left(x\right){,}y^{*}\right)$.~\colldist{} defines the distribution of \(\br{g\br{x},y^*}\) where \(x\sim\basedist\).
This has been defined as the feature-label strategy in prior work.
The signal $g$ also defines the signal set $\collset=\left\{ \collfn\left(x\right)\vert x\in\cX\right\} $.
For any distribution \(\dist\) over \(X{\times} Y\), the Bayes optimal classifier on  \(\dist\), which we denote as $f_\dist$, is defined as
\begin{equation}
    f_\dist\left(x\right)=\arg\max_{y\in Y}\dist\left(Y=y\mid X=x\right).
\end{equation}%

For Bayes (sub-)optimal classifiers,~\citet{hardt_algorithmic_2023} identify four properties that affect the success of a collective action:

\begin{itemize}

\item The fractional size of the collective \strength~(see~\Cref{eq:coll-dist})  gives the collective greater statistical power. The larger the size, the higher the success.

\item 
The uniqueness \uniq{} of the signal.
A signal is \uniq{}-unique if $\basedist\left(\collset\right)\leq\uniq$.
Informally, \uniq{} is the measure of the codomain of the collective transformation $g$ under the probability measure of the base distribution \basedist{}. The more unique the signal~(smaller~\uniq{}), the easier it is to associate the signal to $y^{*}$, leading to higher success.

\item The sub-optimality gap of the signal
is defined as $\displaystyle\subgap{=}\max_{x\in\collset}\max_{y\in \cY}\basedist\br{y{\mid} x}{-}\basedist\br{y^{*}{\mid} x}$. 
It measures the extent to which the collective competes with signals already present in \basedist.
The smaller the sub-optimality gap, the higher the chances of success.

\item The sub-optimality \opti{} of a learned classifier relays how close a classifier is to the Bayes optimal.
It is defined as the smallest total variation (TV) distance between \mixdist{} and a distribution on which the learned classifier is actually Bayes optimal. 
\end{itemize}
Using the above four properties, they derive the following lower bound on the success of the collective.

\begin{thm}[Theorem 1 in~\citet{hardt_algorithmic_2023}]\label{thm:hardt-original}
    Given the mixture distribution \mixdist{} and the feature-label strategy for planting a signal, the success is lower bounded by
    \begin{equation}\label{eq:hardt-ineq}
    \success{\strength} \geq 1 - \br{\frac{1-\strength}{\strength}}\subopt\cdot\uniq - \frac{\opti}{1-\opti},
    \end{equation}
    where \strength{}, \uniq{}, \subgap{}, and \opti{} are, respectively, the size, uniqueness, sub-optimality gap for $y^{*}$ in the base distribution, and sub-optimality of the learned classifier on \mixdist{}.
\end{thm}
While they are sufficient to characterise the success of the collective for a Bayes (sub-)optimal learner, various practically deployed algorithms show different behaviours, as is discussed in \cref{sec:eff-size-val-control,sec:oblivious}.

\subsection{Distributionally Robust Optimisation}
\label{sec:dro-formal}

In~\Cref{sec:eff-size-val-control}, we inspect collective action on a set of learning algorithms that target \textit{Distributionally-Robust} objectives~\citep{delage_distributionally_2010,sagawa_distributionally_2020,duchi_learning_2021}. 
Intuitively, these algorithms aim to learn classifiers that perform equally well on a set of distributions as opposed to any single one.
Formally, they minimise the following objective

\begin{equation}
    \cR_{\mathrm{dro}}(\theta)\coloneqq\sup_{q\in\cQ_{p}}\EE_{q}\left[\ell\left(g_{\theta}(x),y\right)\right],
\end{equation}
where $\cQ_p$ is an \textit{uncertainty set} of distributions close to $p$ over which we want to control the risk, $\ell$ is the loss function, and $g_{\theta}$ is a function with parameters $\theta$.
There are many possible definitions of the uncertainty set that the algorithm is %
controlling for.
One possible choice is an $f$-divergence\footnote{~\Cref{ap:fdiv} in the appendix.} ball~\citep{ali_general_1966,csiszar_information-type_1967}.

\begin{equation}
    \label{eq:dro-uncert}
   \cQ_{p}=\left\{ q\ll p\mid\cD_{f}\left(q\mid\mid p\right)\leq\delta\right\},
\end{equation}
where $\delta$ is the radius according to $\cD_{f}$.\footnote{The notation \(q\ll p\) means \(q\) is absolutely continuous with respect to \(p\).} An often cited use-case of DRO algorithms is to protect the performance of small subgroups~\citep{hashimoto_fairness_2018}.

Consider a set of subgroups or sub-populations denoted as $\cA$ and let the observed distribution, $p$, arise as a mixture over the subgroups with distributions $p_a$ as $p = \sum_{a\in \cA }\alpha_a p_a$ so the uncertainty set then becomes
\begin{equation}
   \cQ_p = \left\{\sum_{a\in \cA }\beta_a p_a  \mid \sum_{a \in \cA} \beta_a=1, \beta_a \geq 0 \right\}.
\end{equation}
Under this setting, minimising $\cR_{\mathrm{dro}}(\theta)$ is equivalent to minimising the worst group loss~\citep{sagawa_distributionally_2020}
\begin{equation}
    \cR_{\mathrm{WGL}}\left(\theta\right)\coloneqq\max_{a\in\cA}\EE_{q}\left[\ell\left(g_{\theta}\left(x\right),y\right)\mid A=a\right].
\end{equation}
Thus, DRO algorithms are often employed when there is concern about performance on ``similar'' distributions or on subgroups of the data~\citep{namkoong_stochastic_2016,duchi_variance-based_2019}.
In~\Cref{sec:eff-size-val-control}, we investigate how the success of the collective changes when minimising $\cR_{\mathrm{dro}}(\theta)$ as opposed to performing simple Empirical Risk Minimisation~(ERM).

\section{Effective Size and Validation Control}
\label{sec:eff-size-val-control}

The most intuitive parameter to predict the success of collective action is the fractional size of the collective \strength{}; a larger collective will attain greater success. However, DRO algorithms assign different weights to different samples, rendering \strength{} inappropriate for predicting success. Instead, we introduce a correction to the collective size under a weighted distribution that we denote the effective size \effsize{}, and show that DRO algorithms can yield $\effsize>\strength$.

We experimentally validate this theory on a selection of two-stage re-weighting algorithms, specifically \jtt{}~\citep{liu_just_2021} and \lff{}~\citep{nam_learning_2020}. Our findings, based on synthetic and image datasets, indicate that DRO algorithms can significantly increase collective success, surpassing standard Empirical Risk Minimization (ERM) for the same tasks.

Finally, we turn to iterative re-weighting algorithms, focusing on \dro{}~\citep{levy_large-scale_2020}. Unlike two stage re-weighting algorithms, iterative re-weighting algorithms oscillate between fitting different parts of the data in training, relying on performance on a validation set as a stopping criterion.
When the collective can influence both the training and validation set, we show that this stopping criteria makes them particularly sensitive to the collective's size in the validation set. To demonstrate this we first analyse an abstract theoretical version of \dro{}, varying the degree of use of collective action in the validation set. Finally, we experimentally validate these claims, showing the collective success with \dro{} is very sensitive to the collective proportion in the validation set. 

\subsection{Effective Collective Size}\label{sec:eff-coll-size}

As mentioned, DRO algorithms allow for varying data points to have differing levels of impact on the algorithm by assigning different weights to the training data.
Let $w:\cX\times\cY\rightarrow\mathbb{R}$ be a mapping from a sample to its weight.
Those weights define the weighted distribution
\begin{equation}
    \cP^{\left(w\right)}\left(X=x,Y=y\right)=\frac{w\left(x,y\right)\cP\left(X=x,Y=y\right)}{\EE_{\cP}\left[w\left(x,y\right)\right]}.
\end{equation}

Within the context of collective action, this weighting can effectively boost or diminish the influence of the collective. To capture this change of impact, we introduce the following notion of effective collective size $\effsize\left(w\right)$.

\begin{defn}
For a distribution $\cP^{\left(w\right)}$ where samples are up-weighted according to their covariates by $w(x)$ we define the \textit{effective collective size} as
\begin{equation}\label{eq:effective_size}\strength_{\text{eff}}\left(w\right)=\frac{\bE_{x,y\sim\cP}\left[w\left(x,y\right)\mathbbm{1}\left\{ (x,y)\text{ is in the collective}\right\} \right]}{\bE_{x,y\sim\cP}\left[w(x,y)\right]}.
\end{equation}
\end{defn}
Note that if $w\br{x}{=}1$ for all $x$, then $\effsize{=}\strength$.
Now, since \effsize{} is the collective size under the weighted distribution, then bounding the success for this algorithm, akin to \Cref{thm:hardt-original}, requires adding a corrective term $c$ for the non-weighted distribution, giving
\begin{equation}
    S\left(\strength\right)\geq1-\left(\frac{1-\effsize}{\effsize}\right)\left(\subopt\cdot\uniq+c\right)-\frac{\opti}{1-\opti},
\end{equation}
with the corrective term being
\begin{equation}
 c=\mathbb{E}_{x\sim\cP^{*}}\left[\frac{\Delta_{x}^{w}\cP_{0}^{\left(w\right)}\left(x\right)}{\left(\cP^{*}\right)^{\left(w\right)}\left(x\right)}-\frac{\Delta_{x}\cP_{0}\left(x\right)}{\cP\left(x\right)}\right],
\end{equation}
where $\Delta_{x}^{w}=\max_{y\in Y}\left(\cP^{\left(w\right)}\left(y\mid x\right)-\cP^{\left(w\right)}\left(y^{*}\mid x\right)\right)$ and $\Delta_{x}$ is defined the same but for \basedist{}. Formal proof and definitions are in the appendix under \Cref{prop:corr-c}.
Under certain circumstances we can have $c\leq 0$, implying a collective success greater than that guaranteed by \Cref{thm:hardt-original} for $\alpha=\effsize$. We provide such an example where $c\leq 0$ and $\effsize\geq \alpha$ in Appendix \ref{ap:alpha_eff}.

We now provide theoretical results, showing that DRO algorithms can increase the effective collective size \effsize{}. Firstly, for any algorithm that targets a DRO defined by an $f$-divergence, we can say the following:

\begin{restatable}{proposition}{drothm}\label{prop:DRo_effective_collective}
For a mixture distribution \mixdist{}, let $\cQ_{\mixdist}$ be the set of distribution in a ball of radius \(\delta\) around \mixdist{} as defined in \cref{eq:dro-uncert}. Then $\cQ_{\mixdist}$ contains a distribution $\cP_{\alpha_{\text{eff}}}$ with effective collective size
\begin{equation}
    \effsize=\strength+\frac{\delta}{\cD_{f}\left(\colldist\Vert\mixdist\right)}.
\end{equation}
\end{restatable}

The proof can be found in \cref{app:theory}. The $\delta$ parameter in this case is the radius of the ball that the algorithm is controlling performance over.
It is either explicitly chosen, or implicitly defined by the hyper-parameters of the algorithm. This proposition tells us that if the algorithm is optimising against a wide range of distributions, this range will include a mixture distribution with a higher \effsize{}.

Now, we turn to analysing two-stage algorithms. In the first phase, these algorithms train a weak classifier, for example with early stopping or strong regularisation.
Then, all samples in the error set of this classifier are up-weighted by a factor of $\lambda$ for the second and final stage of training. This  ensures an algorithm has good performance against the worst case subgroups in the original data. The following characterises the effective collective size for these algorithms.
\begin{restatable}{proposition}{jttthm}\label{prop:Effective_collective_JTT}[Effective Collective Size of JTT~\citep{liu_just_2021}]
    For JTT trained on \mixdist{}, let $\lambda$ be the up-weighting parameter, $f$ be the classifier learned in the first phase and define
    \begin{equation} \label{eq:jtt_probs}
        \begin{aligned}P_{E} & \coloneqq\mixdist\left[f\left(X\right)\neq Y\right]~\text{and}\\
        P_{E\mid C} & \coloneqq\mixdist\left[f\left(X\right)\neq Y\mid\left(X,Y\right)\text{ in the collective}\right].
        \end{aligned}
    \end{equation}
Then, the effective collective size is given by
\begin{equation}
    \effsize=\strength\frac{\lambda P_{E\mid C}+\left(1-P_{E\mid C}\right)}{\lambda P_{E}+\left(1-P_{E}\right)}.
\end{equation}
\end{restatable}

\begin{figure*}[t]\centering
\begin{subfigure}[t]{0.21\linewidth}
        \centering
        \includegraphics[width=\linewidth]{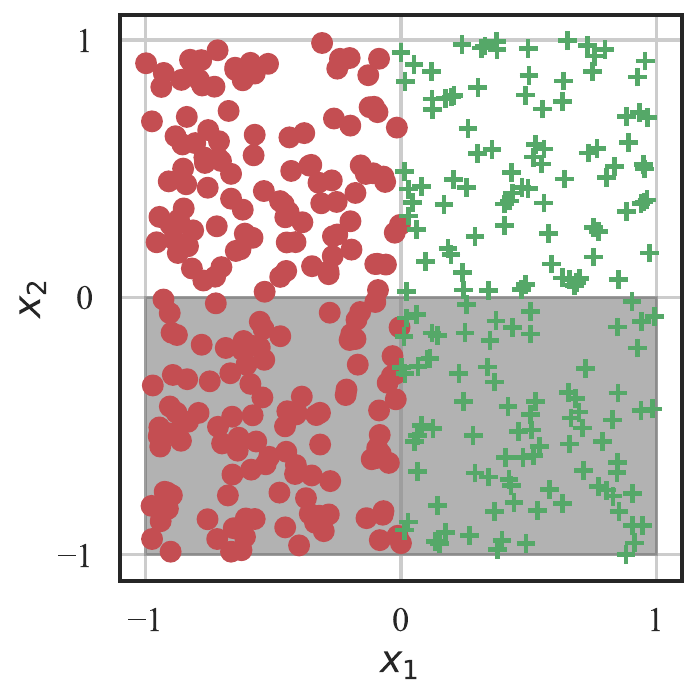}
        \caption{Synthetic 2D dataset}
        \label{fig:example_2d}
    \end{subfigure}
    \begin{subfigure}[t]{0.38\linewidth}
        \centering
        \includegraphics[width=\linewidth]{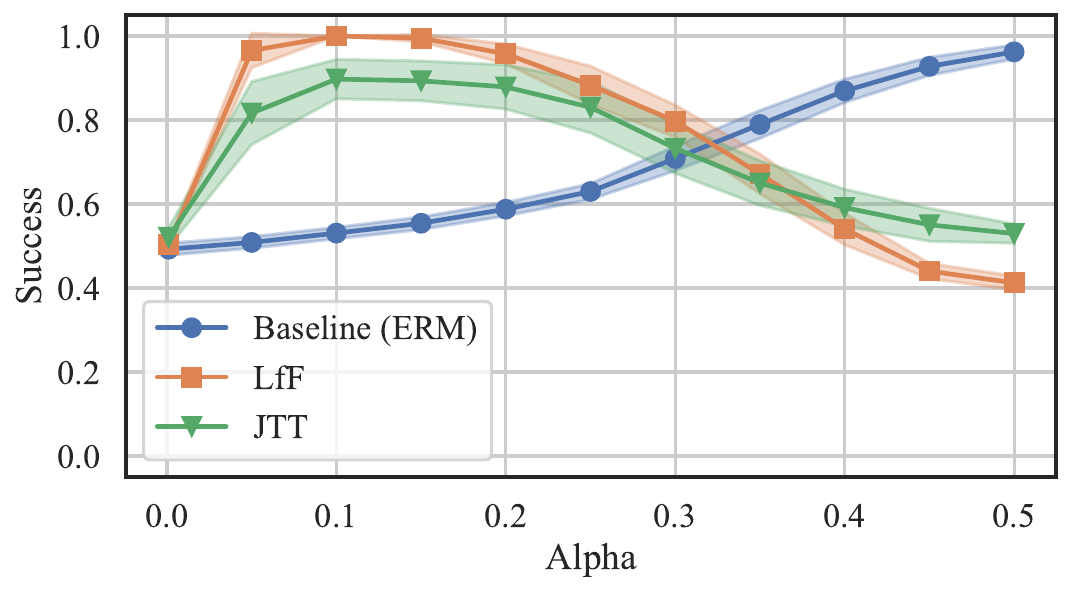}
        \caption{Success on synthetic 2D dataset}
        \label{fig:success_2d}
    \end{subfigure}
    \begin{subfigure}[t]{0.38\linewidth}
        \centering
        \includegraphics[width=\linewidth]{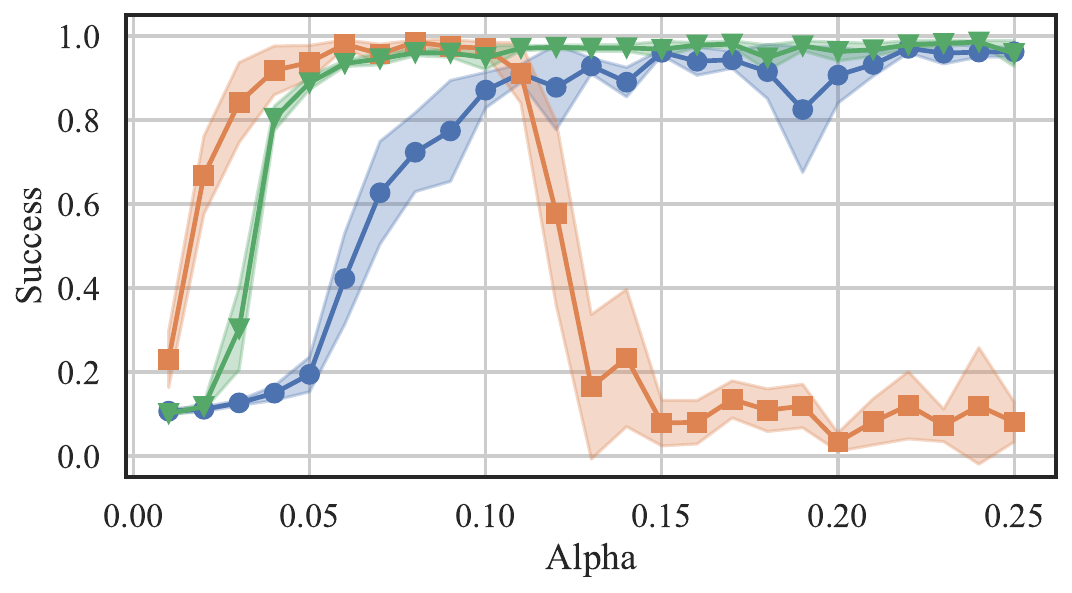}
        \caption{Success on \cifar{}}
        \label{fig:success_cifar}
    \end{subfigure}
    \caption{Success with DRO algorithms. \textbf{(a)} An example of the 2D dataset. The color of each point represents its label, and the grey rectangle is the co-domain of the collective signal. \textbf{(b-c)} The success of a collective of different sizes \strength{} when trained with ERM (blue circles), JTT (orange squares) and \lff{}(green triangles) on a synthetic 2D and CIFAR-10.}
    \end{figure*}

\begin{figure}[t]
    \centering %
    \includegraphics[width=0.45\linewidth]{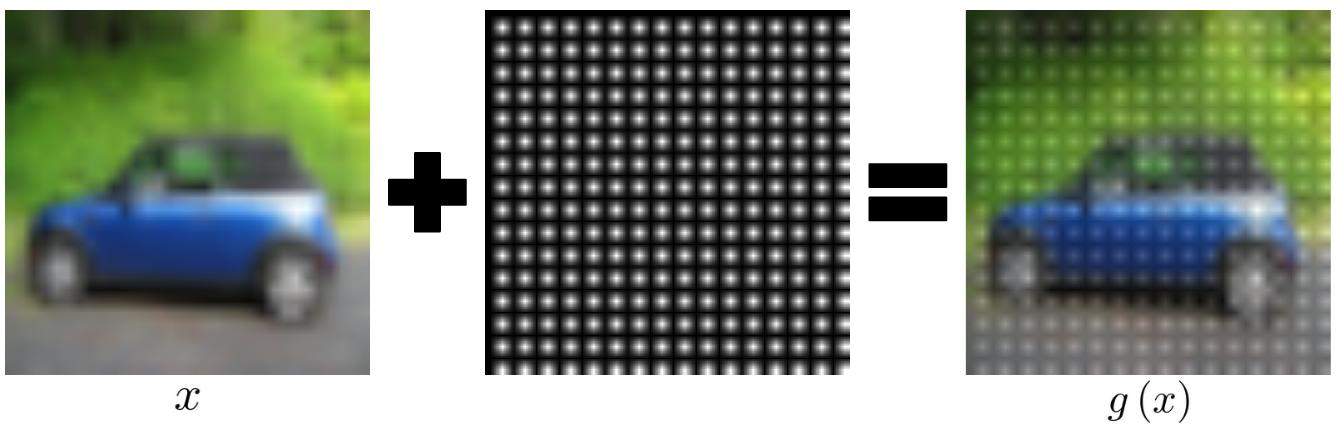}
    \caption{Image transformation used by the collective. The effect of the signal is exaggerated for visualisation purposes and in practice it is invisible to the human eye.}
    \label{fig:cifar_transform}
\end{figure}
The proof can be found in \cref{app:theory}.
This proposition demonstrates that if the first stage classifier $f$ is more likely to make errors on the collective samples than on random samples ($P_{E\mid C} {>} P_{E}$), this leads to $\alpha_{\text{eff}} {>} \alpha$.
This means that if the collective distribution represents a particularly challenging subgroup of the dataset, it will be up-weighted to have a much larger effect on the final output of these algorithms, which should lead to higher collective success.

\subsection{Experiments Results for Two Stage Algorithms}

\begin{figure*}[t]\centering
    \begin{subfigure}[t]{0.49\linewidth}
        \centering
    \includegraphics[width=\linewidth]{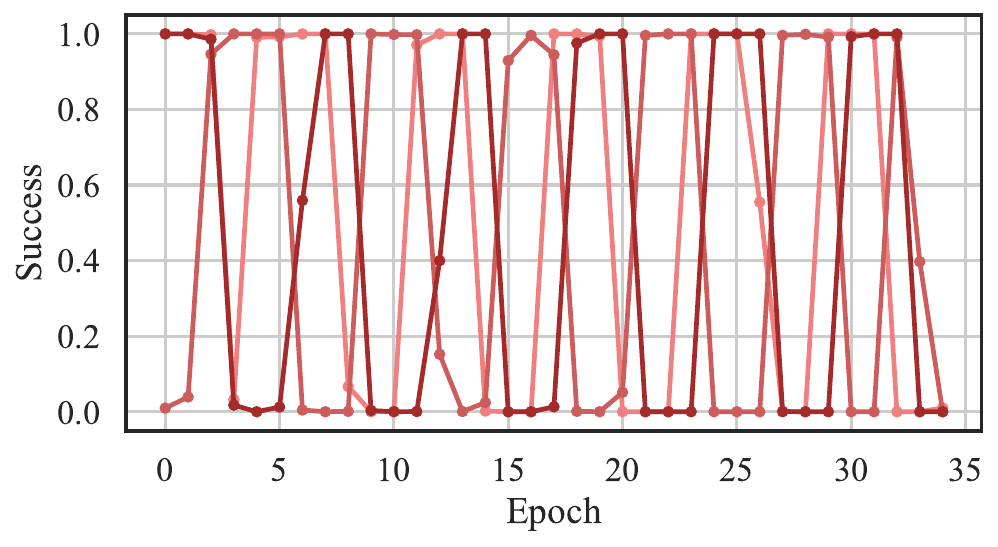}
    \caption{Success against time}
    \label{fig:dro_oscillations}
    \end{subfigure}
    \begin{subfigure}[t]{0.49\linewidth}
        \centering
    \includegraphics[width=\linewidth]{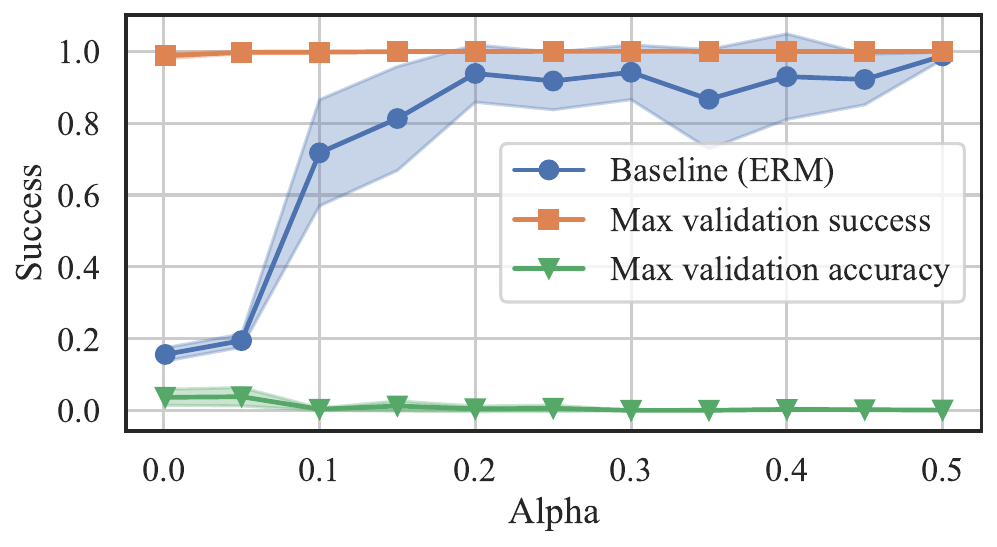}
    \caption{Success for different stopping conditions}
    \label{fig:dro_waterbirds}
    \end{subfigure}
    \caption{Success sensitivity to stopping condition when using \dro{} on the Waterbirds dataset with a collective of size $\strength=0.3$. \textbf{(a)} The success of the collective after every epoch of training. Each shade represents a differently initialised training. The rapid and sharp oscillations show how drastic it is to stop training at the right time. \textbf{(b)} The success achieved by different \strength{} for 2 different stopping conditions. The baseline (ERM) is shown in blue circles. Stopping at maximum accuracy in a validation set with no collective action is shown in the green triangles. Stopping at maximum success on the validation set is shown in orange triangles. When the firm is trying to maximize general accuracy, the collective has no success.}
    \end{figure*}
We experimentally validate the above theory, showing that a collective can be more successful against \jtt{} and \lff{} compared to \erm{} on a synthetic 2D dataset and \cifar{}~\citep{krizhevsky_learning_2009}. These algorithms are explained in \cref{ap:jtt,ap:lff} and the  synthetic 2D dataset comprises points sampled i.i.d. from a uniform distribution on $\left(-1,1\right)^{2}$, labeled by the sign of their first coordinate $x_1$ (\cref{fig:example_2d}).
The collective wants the points with a negative $x_2$ to be labeled $y^*$ and uses the strategy $\left\{ \left(x_{1},x_{2}\right),y\right\} {\rightarrow}\left\{ \left(x_{1},-\left|x_{2}\right|\right),y^{*}\right\}$.
We also consider the multi-class classification problem of~\cifar{}.
The collective transformation \collfn{}, in the pixel space of integer values from 0 to 255, adds a perturbation of magnitude 2 to the value of every second pixel in every second row (\cref{fig:cifar_transform}).
This transform is virtually invisible to the human eye.
Technical details can be found in \cref{sec:experiment_details}.

The results on the 2D dataset (\cref{fig:success_2d}) and on \cifar{} (\cref{fig:success_cifar}) show that for small \strength{}, a collective achieves higher success in \jtt{} and \lff{} than with the same \strength{} in \erm{}.
\cref{prop:Effective_collective_JTT} predicts this behavior for \jtt{}.
When \strength{} is small, the data bias causes \erm{} to misclassify collective members.
This inaccuracy increases the collective population in the error set of first phase.
This leads to a higher \(P_{E\mid C}\) relatively to \(P_{E}\) (\cref{eq:jtt_probs}), which consequently leads to $\effsize>\strength$.
This is no longer the case when \strength{} is large enough.
As \strength{} rises, the accuracy of \erm{} on the collective increases, and the collective membership in the error set decreases.
As a result, $P_{E\mid C}<P_{E}$ and $\effsize<\strength$, lowering the success.
This effect starts at $\strength\approx0.2$ on the 2D dataset and at $\strength\approx0.1$ on \cifar{}.

Intuitively, the goal of \jtt{} and \lff{} is to empower weak groups.
Accordingly, these algorithms give a small collective more statistical power, which grants the collective a higher success than they would achieve in \erm{}.
When the collective is large, these algorithms will take the power away, lowering the success.
This teaches us that a large collective should alter its strategy in order to maximize success.

\subsection{Iterative Re-weighting Algorithms and Collective Action on Validation Sets}
\label{sec:coll-pow-val}

We now turn to analyse algorithms that iteratively re-weight, focusing on \dro{}.
For these algorithms, the effective distribution at each step is chosen adversarially from the uncertainty set of distributions as discussed in Section \ref{sec:dro-formal}.
We explain the \dro{} algorithm in \cref{ap:dro} in the appendix.
This causes the algorithm to cycle between fitting different parts of the distribution, terminating only when a high enough accuracy is reached on some validation set. 
As the validation set plays an important role in deciding when the algorithm terminates, we consider how varying \strength{} in the validation set can affect the collective success.

\paragraph{Theory} In order to theoretically analyse the effect of the validation set on the final success of the collective, we look at an idealised version of a iterative re-weighting DRO algorithm.
This ideal algorithm computes a sequence of classifiers $\cF$, where each classifier $f^{\left(i\right)}$ is a Bayes optimal classifier for a distribution on which the previous classifier $f^{\left(i-1\right)}$ has the maximum error, starting with \basedist{}.
The algorithm outputs the classifier $f{\in}\cF$ that has the highest accuracy on the validation distribution affected by collective action, which is given by $\cP_V = \beta P^* +(1-\beta) \basedist$, where $\beta$ is the collective size in the validation set.
A precise form of this algorithm is given in~\Cref{alg:abstract_cw} in the appendix.
\begin{restatable}{proposition}{drovalthm}\label{prop:dro_val_diff}
    Let $f$ be the output of \cref{alg:abstract_cw}, and $f_{\mixdist}$ be the Bayes optimal classifier on the mixture distribution \mixdist{}.
    Then we have that the success $S_f$ and $S_{f_{\mixdist}}$ with $f$ and $f_{\mixdist}$, respectively, relate as
\begin{equation}
    \displaystyle S_{f}-S_{f_{\mixdist}}{}\geq\frac{\cP_{V}\left[f\left(X\right)=Y\right]-\cP_{V}\left[f_{\mixdist}\left(X\right)=Y\right]}{\beta-\alpha}.
\end{equation}
\end{restatable}
The proof is in \cref{app:theory}.
If $\beta$ is close to $\alpha$, $f_{\mixdist}$ must still be Bayes optimal on $\cP_V$ and so we have that $S_{f}$ is not provably greater than $S_{f_{\mixdist}}$.
However, as $\beta$ increases, $f$ will have better accuracy on the validation set than $f_{\mixdist}$ that was trained on the train set.
In such a case, $\cP_{V}\left[f\left(X\right){=}Y\right]{>}\cP_{V}\left[f_{\mixdist}\left(X\right){=}Y\right]$, which leads to a strictly higher collective success from our idealised \dro{} algorithm when compared to the Bayes optimal classifier on the training distribution.

\paragraph{Experimental Results} 

\cref{fig:dro_oscillations} demonstrates the oscillating success after every epoch when training \dro{} on the Waterbirds dataset~\cite{sagawa_distributionally_2020} with $\alpha{=}0.3$.
These oscillations are due to the search over the distributions space $\cQ_{\mixdist}$(\cref{eq:dro-uncert}). 
\cref{prop:DRo_effective_collective} predicts that this space contains a distribution where the collective is stronger than in the observed distribution.
The peaks of the oscillations suggests that \dro{} was able to find such a distribution.
The frequency and amplitude of the oscillations show that the resulting success is sensitive to when training stops, varying from minimal to maximal success.

Realistically, training stops according to some condition on a validation set.
For example, the firm can choose to stop training at the iteration that achieves the highest accuracy on the validation set.
In \cref{fig:dro_waterbirds} we compare the success on ERM with the success on \dro{} with different conditions on a collective-free validation set: maximum accuracy and maximum collective success.
This comparison shows that when there is no collective action in the validation set, maximum validation accuracy cancels possible success.
The reason that the success is 0 is because there is no collective signal in the validation set.

\begin{figure*}[t]\centering
\begin{subfigure}[t]{0.24\linewidth}
        \centering
        \includegraphics[width=\linewidth]{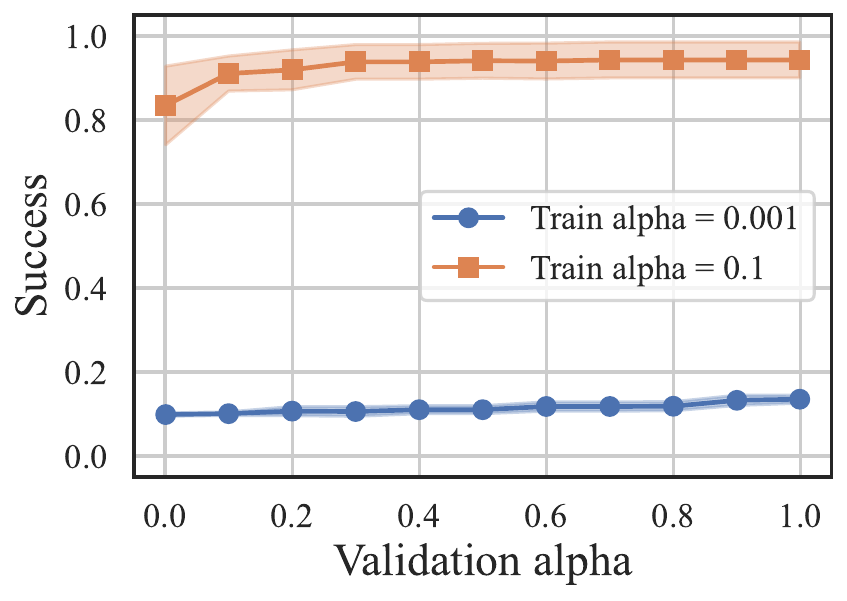}
        \caption{ERM on \cifar{}}
        \label{fig:val_erm_cifar}
    \end{subfigure}
    \begin{subfigure}[t]{0.25\linewidth}
        \centering
        \includegraphics[width=\linewidth]{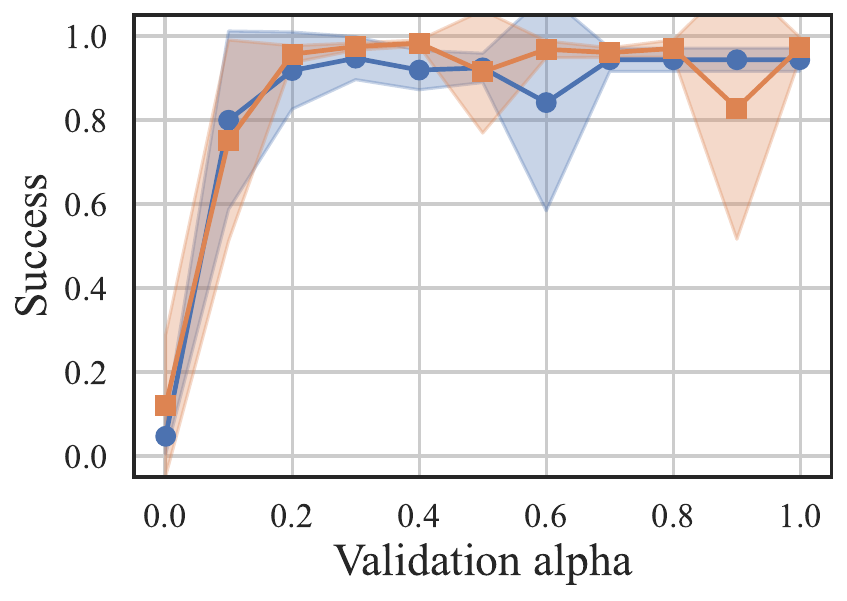}
        \caption{\dro{} on \cifar{}}
        \label{fig:val_dro_cifar}
    \end{subfigure}
    \begin{subfigure}[t]{0.24\linewidth}
        \centering
        \includegraphics[width=\linewidth]{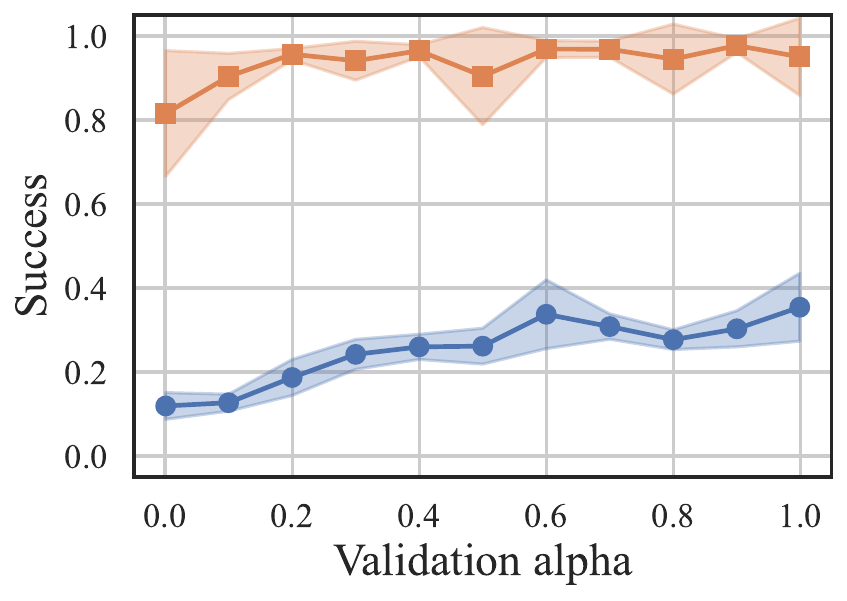}
        \caption{ERM on Waterbirds}
        \label{fig:val_erm_birds}
    \end{subfigure}
    \begin{subfigure}[t]{0.24\linewidth}
        \centering
        \includegraphics[width=\linewidth]{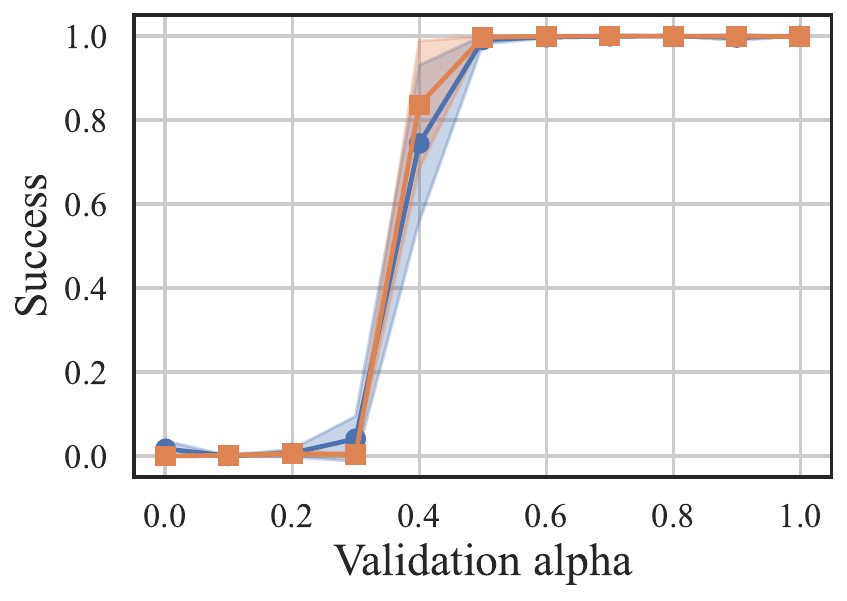}
        \caption{\dro{} on Waterbirds}
        \label{fig:val_dro_birds}
    \end{subfigure}
    
    \caption{Each graph shows the success of different levels on collective action in the validation set $\strength_\text{val}$ when using ERM or \dro{} on \cifar{} and Waterbirds. The blue circles are for a training set with $\strength_\text{train}=0.001$ and the orange squares are for $\strength_\text{train}=0.1$. ERM is almost not affected from $\strength_\text{val}$, but for \dro{} $\strength_\text{val}$ is crucial.}
    \end{figure*}

If the collective is able to affect the validation set, then increasing its proportion in the validation set will increase success, as suggested in \cref{prop:dro_val_diff}.
We show this experimentally by applying collective action in both  training and validation set of the \cifar{} and Waterbirds datasets.
\cref{fig:val_dro_cifar,fig:val_dro_birds} show that when \dro{} is used, the amount of collective in the validation set has a very large impact on the collective success compared to the collective size in the training set.
In contrast, \cref{fig:val_erm_cifar,fig:val_erm_birds} show that success in ERM is more sensitive to the collective size in the training set \strength{} rather than to the size in the validation set.
With $\strength=0.1$ in ERM, the collective achieves almost full success, while when $\strength=0.001$, even with full control of the validations set, the success does not go over $0.4$.

\section{Leveraging Algorithmic Bias}
\label{sec:oblivious}

In the previous section we described how different learning algorithms can modify the effective size of the collective.
However, these results do not provide guidelines for designing a more effective signal \collfn{}. In this section we address this gap by leveraging insights on the biases of the learning algorithm with regard to the properties of the base distribution \basedist{}. This bias deviates the resulting classifier from the Bayes optimal classifier discussed in \citet{hardt_algorithmic_2023}, but in a way that allows the collective to take advantage of.
Essentially, if the learning algorithm overlooks certain signals in the base distribution, it is akin to eliminating these signals from the distribution, rendering them unique.

\subsection{Theoretical Results}

We provide the following theoretical result to support this claim.
First, for an arbitrary distribution \surrdist{} we define the Bayes-optimal classifier on that distribution 
\begin{equation}
    f_{\surrdist}(x) \coloneqq \arg \max_{y \in \cY} \surrdist\left(Y=y|X=x\right),
\end{equation}
and the \surrmixed{}-mixture distribution as
\begin{equation}
    \surrmixed\coloneqq\strength\colldist+\left(1-\strength\right)\surrdist.
\end{equation}

\begin{restatable}{thm}{algbias}\label{thm:oblivious}
    Consider a base distribution \basedist{} and a learning algorithm \alg{} which outputs $\classifier_{\mixdist}$ when learning on \mixdist{}.
     For any given distribution \surrdist{}, we denote the corresponding classifier TV distance as
\begin{equation}\label{eq:obliv}
  \obliv = \mathrm{TV}\left(\colldist\left(X\right){\times} \classifier_{\mixdist}\left(X\right)|\colldist\left(X\right){\times} f_{\surrmixed}\left(X\right)\right).
\end{equation}
Then the collective success of algorithm \alg{} on \mixdist{}, is bounded below as
\begin{equation}
    S\left(\strength\right){\geq}\sup_{\surrdist}\left\{ 1{-}\frac{\obliv}{1-\obliv}{-}\frac{1-\strength}{\strength}\surrunique\surrcompete\right\},
\end{equation}
where $\displaystyle \surrcompete{=}\max_{x\in\cX^{*}}\max_{y\in\cY}{\surrdist}\left(y{\mid} x\right)-{\surrdist}\left(y^{*}{\mid} x\right)$. 
\end{restatable}
The proof is given in \cref{ssec:proof_oblivious}.
Note that the bound in \cref{thm:oblivious} cannot be smaller than the bound in \cref{thm:hardt-original}, since $\surrdist{=}\basedist$ recovers the original bound in \cref{eq:hardt-ineq}.
A toy example where the bound is strictly tighter can be found in \cref{ap:example5}.

\begin{figure*}[t]\centering
\begin{subfigure}[c]{0.35\linewidth}
    \centering
    \includegraphics[width=\linewidth]{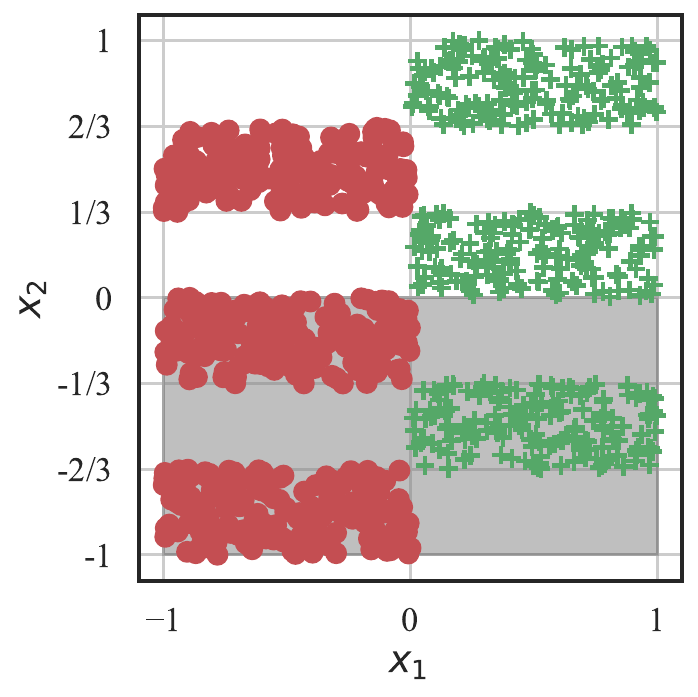}
    \caption{LMS-6 dataset}
    \label{fig:slabs_example}
\end{subfigure}\hfill
\begin{subfigure}[c]{0.55\linewidth}
    \centering
    \includegraphics[width=\linewidth]{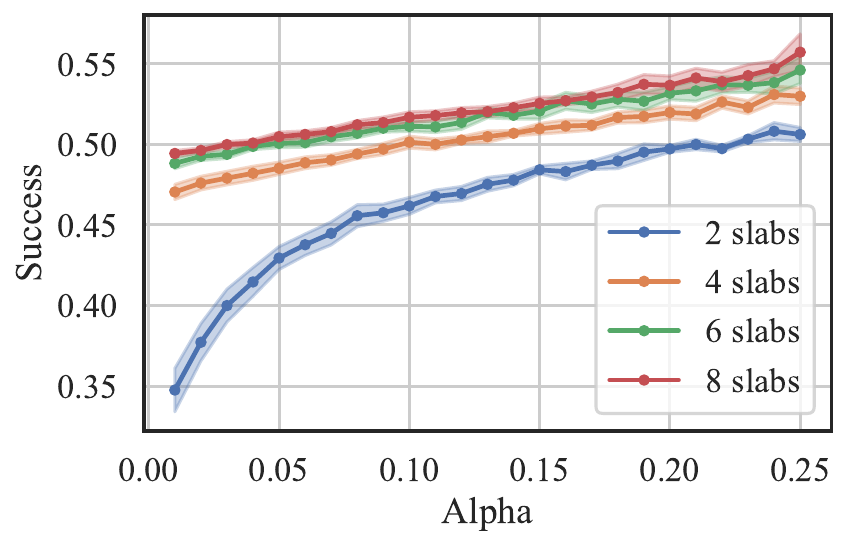}
    \caption{Success on LMS-$k$}
    \label{fig:slabs_success}
\end{subfigure}
\caption{Results on LMS-$k$. \textbf{(a)} an example of LMS-$6$. The color of each point represents its label, the grey rectangle is the codomain of the collective signal. \textbf{(b)} Success over LMS-$k$. Larger complexity ($k$) increases the success.
}
\end{figure*}

Intuitively, if a learning algorithm ignores ``complicated'' features, then one can conceptualise a surrogate distribution \surrdist{} that is devoid of those ``complicated'' features.
The corresponding mixture distribution is then defined as ${\surrmixed{=}\strength\colldist {+} \left(1{-}\strength\right)\surrdist}$.
Within this surrogate distribution, the signal \collfn{} is $\uniq_{\surrdist}$-unique with a smaller $\uniq_{\surrdist}<\uniq$ uniqueness parameter.
As \surrdist{} and \basedist{} share the same features that are relevant to the learned classifier, the Bayes optimal classifier on \surrmixed{} is likely to closely resemble the learnt classifier on the original distribution \mixdist{}.
This similarity effectively causes the signal to be $\uniq_{\surrdist}$-unique on \mixdist{} as well.

\subsection{Experimental Results With Simplicity Bias}
In the experiments, we focus on SGD, which is not only ubiquitous, but also has a bias towards ``simple'' features~\citep{shah_pitfalls_2020}. 
We demonstrate our theory with three different experiments, each showing a different approach for constructing the collective signal \collfn{}.
In all these examples, we train the models using stochastic gradient descent (SGD) (more details in \cref{sec:experiment_details}), leveraging its known preference towards learning simpler features first.

\paragraph{Collective action on a complex feature}

The first approach we explore involves the collective embedding its signal within a complex feature.
We illustrate this approach on a dataset similar to LMS-$k$ from \cite{shah_pitfalls_2020}, shown in \cref{fig:slabs_example}.
In this dataset we consider a two-dimensional binary classification problem with variables \(x_1,x_2\) representing the two dimensions. 
The dataset comprises two blocks along $x_1$, and $k$ blocks along $x_2$.
The classification label is primarily determined by a single threshold function on $x_1$, but can also be inferred using multiple thresholds on $x_2$.
As noted by \citet{shah_pitfalls_2020}, with an increasing number of blocks, models trained using SGD tend to increasingly disregard the $x_2$ feature, classifying by $x_1$ alone.

Here, the collective's goal is to classify samples with $x_{2}{<}0$ as $y^{*}{=}1$.~\Cref{fig:slabs_success} shows that the collective is more successful as \(x_2\) becomes increasingly complex with higher values of $k{\in}\left\{2,4,6,8\right\}$.
\Cref{thm:oblivious} captures this intuition: As $k$ grows, an SGD-based algorithm $\classifier{=}\alg\left(\mixdist\right)$ tends to learn the simpler \(x_1\), becoming oblivious to variations in \(x_2\).
In turn, \classifier{} becomes more similar to a Bayes optimal classifier on a distribution \surrdist{} which is spurious on $x_2$, giving \surrdist{} a smaller \obliv{}, and therefore higher success for larger $k$.

\paragraph{When simpler feature is less informative}

In the previous example, we observed that when both a simple and a complex feature are fully correlated with the label, the collective's success increases as the gap in the simplicity between these features widens.
However, this is not always reflected in practice where the simpler feature may not fully correlate with the label.

\begin{figure*}[tb]\centering
\begin{subfigure}[t]{0.32\linewidth}
    \centering 
    \raisebox{1.5em}{\includegraphics[width=\linewidth]{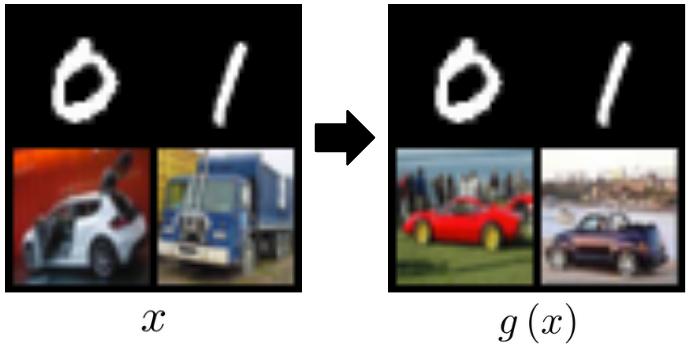}}
    \caption{Collective signal}
    \label{fig:mnistcifar_example}
\end{subfigure}
\begin{subfigure}[t]{0.33\linewidth}
    \centering
    \includegraphics[width=\linewidth]{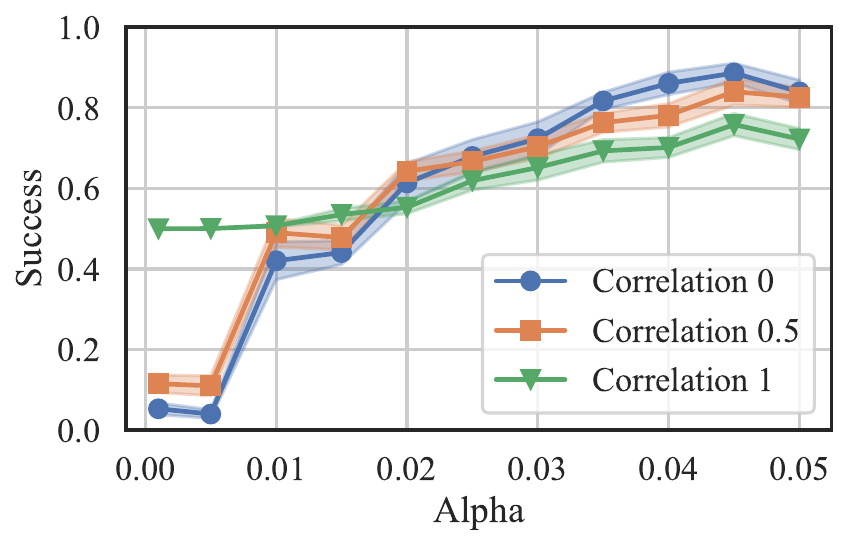}
    \subcaption{On Class Automobile}
    \label{fig:cifarextended_automobile}
\end{subfigure}
\begin{subfigure}[t]{0.33\linewidth}
    \centering
    \includegraphics[width=\linewidth]{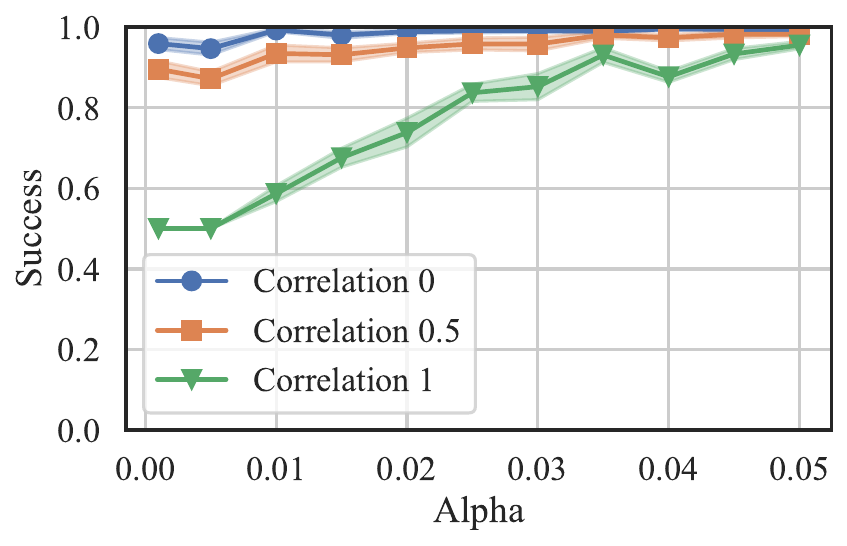}
    \subcaption{On Class Truck}
    \label{fig:cifarextended0_truck}
\end{subfigure}
\caption{Collective success on \textit{correlated} MNIST-CIFAR \textbf{(a)} An automobile and a truck from MNIST-CIFAR and the transformation that changes the CIFAR part to automobile. \textbf{(b-c)} Success with $y^*=\text{truck}$ when the signal is (b) an automobile  or (c) a truck . Blue circles are for no MNIST-to-label correlation, orange square are for 0.5 correlation and green triangles are for full correlation.
}

\label{fig:cifarextended}
    \end{figure*}

One such example is the presence of spurious correlations in the dataset. To demonstrate this, we use another dataset, derived from the MNIST-CIFAR dataset in~\citet{shah_pitfalls_2020}.
Each data point in this dataset comprises a pair of images: one from MNIST (zero or one) and one from CIFAR (truck or automobile), as illustrated in the left side of~\Cref{fig:mnistcifar_example}.
The CIFAR image determines the label.
We then adjust the correlation level between the MNIST image and the label.
A correlation of one implies that automobiles always pair with MNIST zero, and trucks with MNIST one.
A zero correlation indicates random pairing of MNIST digits, and a 0.5 correlation suggests that the MNIST image is randomly sampled half of the time and correlates with the label for the other half.
In this experiment, the collective's strategy involves embedding a signal in the CIFAR component of the dataset, representing either an automobile or a truck, with the designated label \(y^*{=}\text{truck}\).\footnote{Embedding signals in the complex feature is not necessarily the optimal approach in scenarios where the simpler feature is not predictive of the label. }
\cref{fig:mnistcifar_example} shows a transformation using automobile pictures. The full dataset comprises \(5000\) images of trucks and automobiles, but the collective is restricted to only plant a randomly selected subset of a \(100\) of these images.

Our results in \cref{fig:cifarextended_automobile,fig:cifarextended0_truck} show that as the correlation decreases, the success increases.
The underlying intuition is that with higher correlation the simpler MNIST part alone can yield high training performance. As a result, SGD tends to ignore the CIFAR part, where the signal acts.
In contrast, a weaker correlation between the MNIST part and the label necessitates the algorithm's reliance on the CIFAR part, leading it to also learn the collective signal embedded therein.
Note that for small \strength{} in \cref{fig:cifarextended_automobile}, the success with non-zero correlation is low because the collective is labeling automobiles as $y^*{=}\text{truck}$, resulting in competing signals.

This effect can be explained by~\Cref{thm:oblivious} as follows. 
First, note that unlike the analysis in the previous section, in this case it is not possible to design \surrdist{} that is significantly different from \basedist{} in the complex feature~(CIFAR) while maintaining a small \(\obliv\).
This difficulty arises because with weak correlation between the MNIST part and the label, the CIFAR part becomes the primary source of the label information.
Consequently, \surrdist{} must closely resemble \basedist{} in the CIFAR part.
Instead by choosing the surrogate distribution \surrdist{} to be one where the MNIST part contains relatively small amount of information about the label, we can minimise \(\obliv\) while also keeping $\surrunique\surrcompete$ small.
As the correlation increases, this will not be possible and we see a drop in the success of the collective.

\begin{figure*}[t]\centering
\begin{subfigure}[t]{0.35\linewidth}
    \centering
    \includegraphics[width=\linewidth]{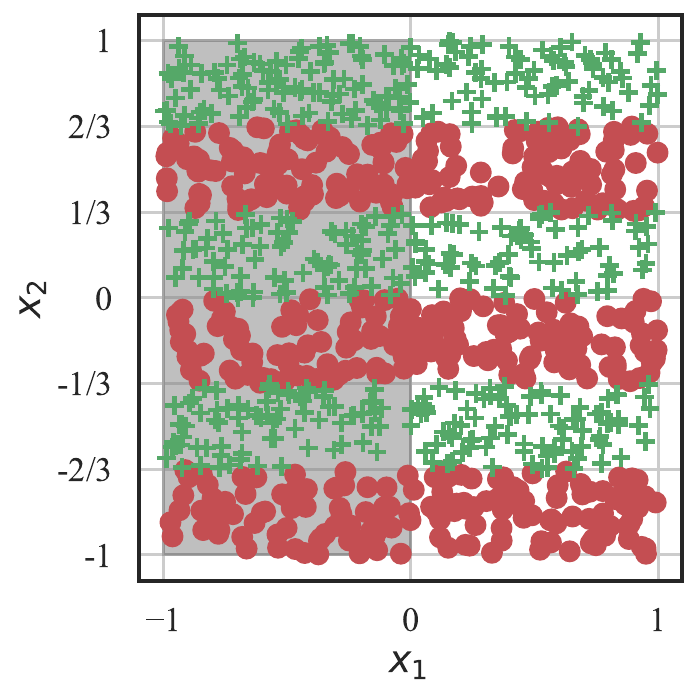}
    \caption{6-strips dataset}
    \label{fig:strips_example}
\end{subfigure}\hfill
\begin{subfigure}[t]{0.55\linewidth}
    \centering
    \includegraphics[width=\linewidth]{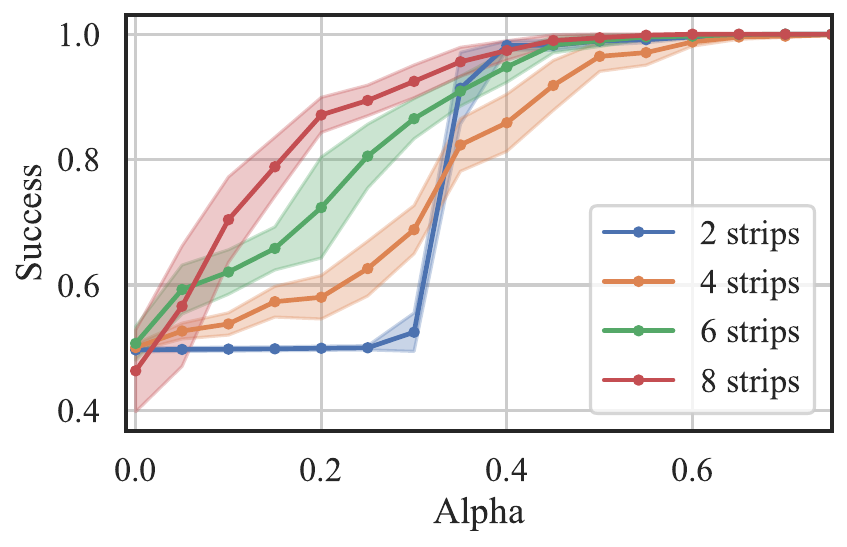}
    \caption{Success of on $k$-strips}
    \label{fig:simplicity_strips}
\end{subfigure}
\caption{Results on $k$-strips. \textbf{(a)} Example of $6$-strips. The color of each point represents its label, the grey rectangle is the codomain of the collective signal. \textbf{(b)} Success for different \strength{} and $k$.}
\end{figure*}

\paragraph{Collective action on simpler feature}

Finally, we demonstrate that when the simpler feature is uncorrelated with the label, the more effective approach is to simply plant the signal in the simpler feature.
This is especially true for algorithms that exhibit simplicity bias and use early stopping or strong regularisation, which are common practices in machine learning.
Intuitively, if the learning algorithm on \mixdist{} is stopped early, it is unlikely to have learned all the correlations present in the data, resulting in a sub-optimal model.
However, it will have captured more of the simpler feature than the complex feature.
Thus, if the collective's signal aligns with the simpler feature, this leads to greater success because the sub-optimal aspects of the classifier will be concentrated in areas outside the collective's signal set.

\begin{remark}
    Note that this phenomenon is not captured by~\citet{hardt_algorithmic_2023}, where the sub-optimality \opti{} is always considered to lie within collective's signal set.
\end{remark}

To demonstrate how a collective can gain from planting a signal in the simpler feature, we use a new dataset, named $k$-strips, which is  similar to LMS-$k$, but removes the correlation between \(x_1\) and the label (\cref{fig:strips_example}).
In this experiment, we create the synthetic dataset by sampling points from a uniform distribution on $\br{0,1}^{2}$, labeled by their $x_{2}$ values. The $x_{2}$-axis is divided into $n$ horizontal strips, and a point has a positive label if it's in an even-numbered strip, and a negative label if it's in an odd-numbered strip.

The collective attempts to focus the attention on $x_{1}$ by giving positive labels if $x_{1}<0$.~\Cref{fig:simplicity_strips} shows that with more horizontal strips, the collective attains higher success with a smaller strength \strength{}.
This result is predicted by \Cref{thm:oblivious}: As $k$ grows, an SGD-based early-stopped algorithm on the collective $\alg\left(\mixdist\right)$ will tend to learn the simpler \(x_1\) feature while largely overlooking \(x_{2}\).
If \(X^{\prime}\subset \reals^{2}\) is the part of the domain where $\alg\left(\mixdist\right)$ accurately predicts on \mixdist{}, a distribution \surrdist{} can place the majority of its probability mass on a thin horizontal strip and the remaining probability mass on \(X^{\prime}\). This will be sufficient to both get a small \(\obliv\) due to the small mass on \(X^{\prime}\), and a small $\surrunique\surrcompete$ due to only small part of \(x_2\) having any probability mass under \surrdist{}.
This leads to a higher lower bound in~\Cref{thm:oblivious}, thereby explaining the observations in~\Cref{fig:simplicity_strips}.

\section{Conclusion and Future Work}
In this work we conducted a theoretical and empirical study on how the choice of learning algorithms affects the success of a collective action to plant a signal.
We presented various approaches a collective can take to design their signal when they have knowledge about the learning algorithm.
In particular, these approaches include maintaining a small collective size against algorithms like~\jtt{} and~\lff{}, influencing the validation set against~\dro{}, and changing the complexity of the signal against~SGD.
While we have focused on these three algorithms, these phenomena are relevant for several popular algorithms.
Common algorithms in the topic of domain adaptation~\citep{koh_wilds_2021}, distribution shift~\citep{hendrycks_benchmarking_2019}, and improving fairness~\citep{berk_convex_2017} use some form of DRO algorithms. It would be interesting to consider the impact of other kinds of algorithms used to improve worst group performance including un-supervsied and self-supervised representation learning~\citep{shi2023how} and  fairness-inducing in-processing~\citep{prost2019toward} and post-processing algorithms~\citep{ctifrea2023frapp}. In a similar vein, recent research has seen a surge in algorithms that aim to improve~\emph{safety} of outputs of generative models including techniques like adversarial training~\citep{ziegler2022adversarial}, DPO~\citep{rafailov2024direct}, and RLHF~\citep{christiano2017deep}. It is important to also consider the impact of these algorithms on the success of a well-designed collective. 

Algorithmic biases akin to simplicity bias are also exhibited by SGD training of various state-of-the art deep neural networks. This includes a texture bias for CNN~\citep{hermann_origins_2020}, an in-context bias for language models~\citep{levine_inductive_2022}, and word-order biases in LSTMs and transformers~\citep{white_examining_2021}. 
Our work discusses how these biases can affect the design of the collective signal and how successful they are expected to be. However, several algorithms also display a different kind of biases. For example, differentially private algorithms are unable to learn minority sub-populations~\citep{bagdasaryan2019differential,sanyal2022unfair} and adversarially robust algorithms~\citep{madry2018towards} are used to improve the smoothness of learned models. One direction for future work is to consider the impact of these algorithmic biases on the success of the collective. 

Finally, a third important direction of future research is understanding what level of information and influence is required by the collective to design their signal. For example, our results suggest that improving success against certain DRO algorithms requires less information about the learning algorithm compared to success against SGD style algorithms. However, for iterative re-weighting algorithms like CVar-DRO~\citep{sagawa_distributionally_2020}, it is important to influence both the validation set and the training set. Future work should investigate whether a uniformly randomly selected collective can be as powerful as a strategically chosen collective~(e.g. in adversarial vulnerability~\citep{paleka2023a}). We hope that our work will inspire further research into practical and algorithm-dependent view on collective action.

\section*{Acknowledgements}
The authors thank Alexandru Tifrea for helpful discussions and suggestions. OB is supported by the International Max Planck Research School for Intelligent Systems (IMPRS-IS). JF is funded by the ESRPC.

\bibliographystyle{plainnat}
\bibliography{main_arxiv_ref}

\newpage
\appendix
\onecolumn
\section{Definitions and Algorithms}
\begin{defn} \label{ap:fdiv}
$f$-divergence is a function that measures the difference between two distribution $P$ and $Q$.
For a given convex function $f{:}\left[0,+\infty\right){\rightarrow}\left(-\infty,+\infty\right]$ such that $f\left(x\right)$ is finite for all $x{>}0$, with $f\left(1\right){=}0$ and $\displaystyle f\left(0\right){=}\lim_{t\rightarrow0^{+}}f\left(t\right)$, the $f$-divergence is defined as
\[
D_{f}\left(P\Vert Q\right)=\int_{\Omega}f\left(\frac{\text{d}P}{\text{d}Q}\right)\text{d}Q.
\]    
\end{defn}

\begin{defn}
    \label{ap:jtt}
JJT, as defined by \citet{liu_just_2021}, has two stages.
The first stage trains a classifier by ERM.
The second stage trains a different classifier on a similar dataset, where each sample that was misclassified by the classifier from the first stage, is given a higher weight.
\begin{algorithm}
\begin{flushleft}
    \begin{algorithmic}
    \textbf{Input:} Training set $\mathcal{D}$ and hyperparameters $T$ and $\lambda_{\text{up}}$.
    
    \textbf{Stage one: identification}
    
        1. Train $\hat{f}_{\text{id}}$ on $\mathcal{D}$ for $T$ steps. 

        2. Construct the error set $E$ of training examples misclassified by $\hat{f}_{\text{id}}$.
    
    \textbf{Stage two: upweighting identified points}
        
        3. Construct upsampled dataset $\mathcal{D}_\text{up}$ containing examples in the error set $\lambda_{\text{up}}$ times and all other examples once.
        
        4. Train final model $\hat{f}_{\text{final}}$ on $\mathcal{D}_\text{up}$.
    \end{algorithmic}
    \end{flushleft}
    \caption{JTT training}
    \label{alg:jtt}
\end{algorithm}
\end{defn}
\begin{defn}
    \label{ap:lff}
LfF, as defined by \citet{nam_learning_2020}, simultaneously trains 2 models: a biased classifier $f_{B}$, and a de-biased classifier $f_{D}$.
The biased model is encouraged to learn biases by using a generalized cross entropy (GCE) loss, and the de-biased model is trained by giving more weight to samples that the biased model fails on.
\begin{algorithm}
\caption{Learning from Failure}
\label{alg:name}
\begin{algorithmic}[1]
\STATE \textbf{Input}: $\theta_B, \theta_D$, training set $\mathcal{D}$, learning rate $\eta$, number of iterations $T$
\STATE Initialize two networks $f_B(x; \theta_B)$ and $f_D(x; \theta_D)$.
\FOR{$t = 1, \cdots, T$}
\STATE Draw a mini-batch $\mathcal{B} = \{(x^{(b)}, y^{(b)})\}_{b=1}^{B}$ from $\mathcal{D}$
\STATE Update $f_B(x; \theta_B)$ by $\theta_{B} \gets \theta_{B} - \eta \nabla_{\theta_{B}} \sum_{(x,y)\in\mathcal{B}}\textrm{GCE}(f_{B}(x), y)$.
\STATE Update $f_D(x; \theta_D)$ by $\theta_{D} \gets \theta_{D} - \eta \nabla_{\theta_{D}} \sum_{(x,y)\in\mathcal{B}} \mathcal{W}(x) \cdot \textrm{CE}(f_{D}(x), y)$.
\ENDFOR
\end{algorithmic}
\end{algorithm}
Where CE stands for the cross entropy loss, and GCE with hyperparameter $q$ is defined as
\[
\text{GCE}\left(p\left(f\right),y\right)=\frac{1-\left(f\left(y\right)\right)^{q}}{q},
\]
where $f\left(y\right)$ is the probability of label $y$ after a softmax layer.
The weight $\mathcal{W}$ per sample is defined as
\[
\mathcal{W}\left(x\right)=\frac{\text{CE}\left(f_{B}\left(x\right),y\right)}{\text{CE}\left(f_{B}\left(x\right),y\right)+\text{CE}\left(f_{D}\left(x\right),y\right)}.
\]
\end{defn}
\begin{defn}
    \label{ap:dro}
CVaR-DRO, as defined by \citet{levy_large-scale_2020}, dynamically changes the weights of samples according to their loss at every iteration.
After the loss is computed for every sample in the mini-batch, and the samples with the smallest are ignored (given a 0 weight) in the current update step.
\end{defn}

\section{Theoretical Results}
\label{app:theory}

In this section, we provide theoretical proofs for results stated in the main text.

\subsection[Impact of effective alpha]{Impact of \effsize{}}
\label{ap:alpha_eff}
\begin{proposition}
Suppose we have a set of weights $w: \cX \times \cY \to \mathbb{R}^{+}$, then we have that:
\begin{align*}
    \mixdist^{(w)} = \alpha_{\mathrm{eff}} \left(\colldist\right)^{(w)} +(1-\alpha_{\mathrm{eff}}) \basedist^{(w)}
\end{align*}
\end{proposition}
\begin{proof}
    We have that:
\begin{align*}
    \mathrm{Pr}_{\mixdist^{(w)}}(X=x,Y=y) &= \frac{w(x,y)\left(\alpha \mathrm{Pr}_{\left(\colldist\right)^{(w)}}(X=x,Y=y) + (1-\alpha)\mathrm{Pr}_{\basedist^{(w)}}(X=x,Y=y) \right)}{{\EE_{\mixdist}[w(x,y)]}} \\
    &= \frac{\alpha \EE_{\colldist}[w(x,y)]\mathrm{Pr}_{\left(\colldist\right)^{(w)}}(X=x,Y=y) + (1-\alpha)\EE_{\basedist}[w(x,y)]\mathrm{Pr}_{\basedist^{(w)}}(X=x,Y=y) }{{\EE_{\mixdist}[w(x,y)]}}\\
    &= \frac{\alpha \EE_{\colldist}[w(x,y)]}{{\EE_{\mixdist}[w(x,y)]}}\mathrm{Pr}_{\left(\colldist\right)^{(w)}}(X=x,Y=y) + \frac{(1-\alpha) \EE_{\basedist}[w(x,y)]}{{\EE_{\mixdist}[w(x,y)]}}\mathrm{Pr}_{\basedist^{(w)}}(X=x,Y=y) \\
    &= \alpha_{\mathrm{eff}} \left(\colldist\right)^{(w)} +(1-\alpha_{\mathrm{eff}}) \basedist^{(w)}
\end{align*}
Where the final line follows as we have $\frac{\alpha \EE_{\colldist}[w(x,y)]}{{\EE_{\mixdist}[w(x,y)]}} = \frac{\EE_{\mixdist}[w(x,y) \mathbbm{1}(\text{Sample from collective})]}{{\EE_{\mixdist}[w(x,y)]}}= \alpha_{\mathrm{eff}}$ .
\end{proof}
\begin{corollary}
    For any $x \in \cX$ we have $f(x) = y^*$ if:
\begin{align*}
   \alpha_{\text{eff}} > (1-\alpha_{\text{eff}})\frac{\Delta^w_x\basedist^{(w)}(x)}{(\colldist)^{(w)}(x)} 
\end{align*}
\end{corollary}
\begin{proof}
    This follows from the same argument as \citet{hardt_algorithmic_2023} where we now use the weighted distributions. 
\end{proof}
\begin{proposition}\label{prop:corr-c}
    Let:
\begin{align*}
    c =\EE_{\colldist}\left[ \frac{\Delta^w_x\basedist^{(w)}(x)}{(\colldist)^{(w)}(x)} \right] - \EE_{\colldist}\left[ \frac{\Delta_x\basedist(x)}{\colldist(x)} \right]
\end{align*}
Where $\Delta_x = \max_{y}\left(\basedist(y\mid x)-\basedist(y^*\mid x) \right)$ and $\Delta^w_x = \max_{y}\left(\basedist^{(w)}(y\mid x)-\basedist^{(w)}(y^*\mid x) \right)$.Then we have: 
\begin{align}
     \success{\strength} \geq 1 - \br{\frac{1-\strength_{\mathrm{eff}}}{\strength_{\mathrm{eff}}}} \left(\subopt\cdot\uniq + c \right)- \frac{\opti}{1-\opti},
\end{align}
Moreover, if we have that:
\begin{align*}
      w(x,y^{\prime}) \leq \frac{\Delta_x E_{x,y \sim \mathcal{P}^0}\left[ w(x,y)\right] } {\Delta^w_xE_{x,y \sim \mathcal{P}^\star}\left[ w(x,y)\right] } w(x,y^\star)
\end{align*}
Then $c\leq 0 $ so that:
\begin{align}
     \success{\strength} \geq 1 - \br{\frac{1-\strength_{\mathrm{eff}}}{\strength_{\mathrm{eff}}}} \subopt\cdot\uniq - \frac{\opti}{1-\opti},
\end{align}

\begin{proof}
First, for any $x \in \cX$ we have $f(x) = y^*$ if:
\begin{align*}
   \alpha_{\text{eff}} > (1-\alpha_{\text{eff}})\frac{\Delta^w_x\basedist^{(w)}(x)}{(\colldist)^{(w)}(x)} 
\end{align*}
Now, as in the original proof of \cite{hardt_algorithmic_2023}, we have that if the classifier is Bayes optimal on $\mixdist$:
\begin{align}
    S(\alpha) &\geq 1-\frac{1-\alpha_{\mathrm{eff}}}{\alpha_{\mathrm{eff}}} \EE_{x \sim \colldist} \left[ \frac{\Delta^w_x\basedist^{(w)}(x)}{(\colldist)^{(w)}(x)}  \right] \\
    &=1-\frac{1-\alpha_{\mathrm{eff}}}{\alpha_{\mathrm{eff}}} \EE_{x \sim \colldist} \left[ \frac{\Delta_x\basedist(x)}{\colldist(x)}  \right] + \frac{\left(1-\alpha_{\mathrm{eff}} \right) c}{\alpha_{\mathrm{eff}}} 
\end{align}
Now, we have that $c$ can be written as:
\begin{align}
    c &= \EE_{\colldist}\left[ \frac{\Delta^w_x\basedist^{(w)}(x)}{(\colldist)^{(w)}(x)}  - \frac{\Delta_x\basedist(x)}{\colldist(x)} \right] \\
    &=\EE_{\colldist}\left[ \frac{1}{(\colldist)^{(w)}(x)} {\sum_{y^{\prime}\in \cY} \Delta^w_x \frac{w(x,y^{\prime})}{\EE_{x,y \sim \basedist}\left[ w(x,y)\right] } \basedist(x,y^{\prime})} - \frac{1}{(\colldist)^{(w)}(x)} { \sum_{y^{\prime}\in \cY}  \Delta_x \frac{w(x,y^*)}{\EE_{x,y \sim \colldist}\left[ w(x,y)\right] } \basedist(x,y^{\prime})} \right] \\
    &=\EE_{\colldist}\left[ \frac{1}{(\colldist)^{(w)}(x)} \left({\sum_{y^{\prime}\in \cY} \Delta^w_x \frac{w(x,y^{\prime})}{\EE_{x,y \sim \basedist}\left[ w(x,y)\right] } \basedist(x,y^{\prime})} - { \Delta_x \frac{w(x,y^*)}{\EE_{x,y \sim \colldist}\left[ w(x,y)\right] } \basedist(x,y^{\prime})} \right)\right] \\
    &=\EE_{\colldist}\left[ \frac{1}{(\colldist)^{(w)}(x)} \left({\sum_{y^{\prime}\in \cY} \Delta^w_x \frac{w(x,y^{\prime})}{\EE_{x,y \sim \basedist}\left[ w(x,y)\right] } \basedist(x,y^{\prime})} - { \Delta^w_x \frac{\Delta^w_x w(x,y^*)}{\Delta_x\EE_{x,y \sim \colldist}\left[ w(x,y)\right] } \basedist(x,y^{\prime})} \right)\right] \\
    &=\EE_{\colldist}\left[ \frac{1}{(\colldist)^{(w)}(x)} \left({\sum_{y^{\prime}\in \cY} \Delta^w_x \basedist(x,y^{\prime})} \left( {w(x,y^{\prime})}- { \frac{\Delta_x {\EE_{x,y \sim \basedist}\left[ w(x,y)\right] }}{\Delta^w_x\EE_{x,y \sim \colldist}\left[ w(x,y)\right] } w(x,y^*) } \right)\right)\right] \\
\end{align}
Where this term is negative if $w(x,y^{\prime}) - { \frac{\Delta_x {\EE_{x,y \sim \basedist}\left[ w(x,y)\right] }}{\Delta^w_x\EE_{x,y \sim \colldist}\left[ w(x,y)\right] } w(x,y^*) } \leq 0 $ which happens when $w(x,y^{\prime}) \leq  { \frac{\Delta_x {\EE_{x,y \sim \basedist}\left[ w(x,y)\right] }}{\Delta^w_x\EE_{x,y \sim \colldist}\left[ w(x,y)\right] } w(x,y^*) } $
\end{proof}.

\paragraph{Example} 
If we have  $\frac{\Delta_x}{\Delta^w_x} \geq \lambda$ for all $x$ then if for $w(x,y^*)= \lambda \EE_{x,y \sim \basedist}\left[ w(x,y)\right]$ we have that:
\begin{align*}
    w(x,y^{*}) - { \frac{\Delta_x {\EE_{x,y \sim \basedist}\left[ w(x,y)\right] }}{\Delta^w_x\EE_{x,y \sim \colldist}\left[ w(x,y)\right] } w(x,y^*) } &= \lambda \EE_{x,y \sim \basedist}\left[ w(x,y)\right] - { \frac{\Delta_x {\EE_{x,y \sim \basedist}\left[ w(x,y)\right] }}{\Delta^w_x}} \\
    &\leq 0
\end{align*}
For all other $y^{\prime}$, setting $w(x,y^{\prime}) \leq  { \frac{\Delta_x {\EE_{x,y \sim \basedist}\left[ w(x,y)\right] }}{\Delta^w_x}}$ is sufficient for $w(x,y^{\prime}) - { \frac{\Delta_x {\EE_{x,y \sim \basedist}\left[ w(x,y)\right] }}{\Delta^w_x\EE_{x,y \sim \colldist}\left[ w(x,y)\right] } w(x,y^*) } \leq 0 $. Finally we can see that $\effsize{} = \lambda \alpha$ so setting $\lambda\geq 0$ we have a setting where $c \leq 0$ and $\effsize \geq \alpha$.
\end{proposition}
\subsection[Proofs for Results in Section 3]{Proofs for Results in~\Cref{sec:eff-size-val-control}}
\label{ssec:proof_prop2}

\drothm*
\begin{proof}
Letting:
\begin{equation*}
    \dist=\strength\colldist + \left(1-\strength\right)\basedist,
\end{equation*}
Then we want to consider for what $\lambda$ does the mixture distribution $ \lambda P^* + (1-\lambda) \dist $ lies in $\cQ_{\delta}$. Now we have that:
\begin{align*}
    \cD_f(\lambda \cP^* + (1-\lambda) \dist \mid\mid \dist) &\leq \lambda \cD_f( \cP^* \mid\mid \dist)+ (1-\lambda) \cD_f( \dist \mid\mid \dist) \\
    &=\lambda \cD_f( \cP^* \mid\mid \dist)
\end{align*}
Now, we have that:
\begin{align*}
    \cD_f( \cP^* \mid\mid \dist) &= \cD_f( \cP^* \mid\mid \strength\colldist + \left(1-\strength\right)\basedist) \\
    &= \cD_f( \cP^* \mid\mid \strength\colldist + \left(1-\strength\right)\basedist) \\
    &\leq \alpha \cD_f( \cP^* \mid\mid \cP^*) +(1-\alpha) \cD_f( \cP^* \mid\mid \cP) \\
    &=(1-\alpha) \cD_f( \cP^* \mid\mid \cP)
\end{align*}
Now plugging in the above we have that the following is sufficient to guarantee $\lambda P^* + (1-\lambda) \dist \in \cQ_{\delta}$:
\begin{align*}
    \lambda \leq \frac{\delta}{(1-\alpha) \cD_f( \cP^* \mid\mid \cP)}
\end{align*}
Finally, note the collective proportion in $\lambda \cP^* + (1-\lambda) \dist $ is $\alpha +\lambda - \alpha\lambda$. Plugging in the largest $\lambda$ and collecting terms gives the desired result. 
\end{proof}

\jttthm*
\begin{proof}
    We have that the effective collective size is defined as:
\begin{equation*}
    \alpha_{\text{eff}} = \frac{\bE[w(X) \mathbbm{1} \{ \text{X is in the collective} \}]}{\bE[w(X)]}
\end{equation*}
In this case we have that $\EE[w(X)] = \lambda P_{E} + (1-P_{E})$ and that:
\begin{align*}
   \bE[w(X) \mathbbm{1} \{ \text{X is in the collective}\}] &= P(\text{X is in the collective}) \EE[w(X) \mid \text{X is in the collective})] \\
   &= \alpha \left(\lambda P_{E\mid C} + (1-P_{E\mid C}) \right)
\end{align*}
Inputting these expression into the effective collective size gives the result.

\end{proof}

\paragraph{Validation Control}  For some validation distribution $\cP_V = \beta \cP^* + (1-\beta)\cP_0$ and some minimal acceptable error $\xi$ on the validation set, we  first define an abstract version of \dro{} in~\Cref{alg:abstract_cw}.
\begin{algorithm}
   \caption{Ideal Continuous Reweighting}
   \label{alg:abstract_cw}
\begin{algorithmic}
   \STATE {\bfseries Input:} Validation Distribution $\cP_V = \beta \cP^* + (1-\beta)\cP_0$, Uncertainty ball $\cQ_p$ centred at $\cP = \alpha \cP^* + (1-\alpha)\cP_0$
   \STATE $\displaystyle f_0(x) \leftarrow \arg \max_{y\in \cY} \cP(Y=y \mid X=x)$
    \STATE $t \leftarrow 1$
   \WHILE{$t \leq T $ }
   \STATE $\displaystyle \cP \leftarrow \arg \max_{Q\in\cQ_p} \EE_{Q} \left[ \ell \left(f_{t-1}(x),y\right)\right] $
   \STATE $\displaystyle f_t(x) \leftarrow \arg \max_{y\in \cY} \cP(Y=y \mid X=x)$
   \STATE $t \leftarrow t+1$
   \ENDWHILE
   \STATE $\displaystyle t_{\max} = \arg \max_{t \leq t} \EE_{\cP_V} \left[ \ell \left(f_{t+1}(x),y\right)\right]$
   \STATE \textbf{Return:} $f_{t_{\max}}$
\end{algorithmic}
\end{algorithm}
Then we restate and prove~\Cref{prop:dro_val_diff}.
\drovalthm*
\begin{proof}
This follows as we have:
\begin{align*}
  \mathrm{Pr}_{\cP_V}(f(X)=Y) -\mathrm{Pr}_{\cP_V}(f_{\mixdist}(X)=Y) =\frac{\beta-\alpha}{1-\alpha} \left(S_f-S_{f_{\mixdist}} \right)+ \frac{1-\beta}{1-\alpha} \left(\mathrm{Pr}_{\cP}(f(X)=Y) - \mathrm{Pr}_{\cP}(f_{\mixdist}(X)=Y) \right) 
\end{align*}
But as $f_{\mixdist}$ is Bayes optimal on $\cP$ we have that $\left(\mathrm{Pr}_{\cP}(f(X)=Y) \leq \mathrm{Pr}_{\cP}(f_{\mixdist}(X)=Y) \right)$ which implies that:
\begin{align*}
    \left(\beta-\alpha \right) \left(S_f-S_h \right) \geq \mathrm{Pr}_{\cP_V}(f(X)=Y) -\mathrm{Pr}_{\cP_V}(f_{\mixdist}(X)=Y)
\end{align*}
Re-arranging terms gives the result.
\end{proof}

\subsection[Proofs for results in section 4]{Proofs for Results in~\Cref{sec:oblivious}}
\label{ssec:proof_oblivious}
\algbias*
\begin{proof}

Let $Q$ be any distribution satisfying:
\begin{align*}
  \mathrm{TV}( \cP_{\alpha}(X) \times h_{\cP_\alpha}(X), \cP_{\alpha}(X) \times f_{Q_{\alpha}}(X) ) \leq \obliv
\end{align*}
Now following \citet{hardt_algorithmic_2023}, we have that $f_{Q_{\alpha}}(x)=y$ if:
\begin{align*}
    \alpha > \left(1-\alpha \right) \Delta_{x,Q} \frac{Q(x)}{P^*(x)}
\end{align*}
Where $\Delta_{x,Q} = \max_{y\in \cY} Q(y\mid x) - Q(y^*\mid x)$. 
Therefore, following a similar argument to \citet{hardt_algorithmic_2023}, we have that:
\begin{align*}
    S(\alpha) &= \mathrm{Pr}_{x \sim \cP^*} \left\{ f(x) = y^* \right\} \\
    &\geq \mathrm{Pr}_{x \sim \cP^*} \left\{ \alpha > \left(1-\alpha \right) \Delta_{x,Q} \frac{Q(x)}{P^*(x)} \right\} \\
    &= \EE_{x \sim \cP^*} \left[ \mathbbm{1} \left\{1 - \frac{\left(1-\alpha \right)}{\alpha} \Delta_{x,Q} \frac{Q(x)}{P^*(x)} > 0
    \right\} \right] \\
    &\geq  \EE_{x \sim \cP^*} \left[  1 - \frac{\left(1-\alpha \right)}{\alpha} \Delta_{x,Q} \frac{Q(x)}{P^*(x)}
     \right] \\
     &= 1 - \frac{\left(1-\alpha \right)}{\alpha} \EE_{x \sim \cP^*} \left[  \Delta_{x,Q} \frac{Q(x)}{P^*(x)}
     \right] \\
     & \geq 1 - \frac{\left(1-\alpha \right)}{\alpha} \left(Q(\cX^*)\Delta_Q \right) \\  
\end{align*}
Now the total variation constraint can be added to give that the success under $A(\cP_{\alpha})$ is
\begin{align*}
   S(\alpha) \geq 1 -\frac{\omega}{1-\omega} - \frac{\left(1-\alpha \right)}{\alpha} \left(Q(\cX^*)\Delta_Q \right).
\end{align*}
\end{proof}

\paragraph{Example} An example of where this would hold be when $P_0(y^*\mid x) = 0$, $w(x,y^*) = a$ where $a$ is constant, and $w(x,y^{\prime}) \leq \mathbb{E}_{x,y \sim \mathcal{P}^0}\left[ w(x,y)\right]$ for all $x \in \mathcal{X}^*$. Intuitively, this corresponds to a scenario where the algorithm places lower weight the region on $\cX^*$ than average.

\subsection[Example for Thm. 5]{Example for \Cref{thm:oblivious}} \label{ap:example5}

\begin{proposition}\label{prop:example5}
    There exists a problem setting, defined by a data distribution~\(\basedist\), a biased learning algorithm~\(\alg\), and collective signal \(g\) such that the success obtained with~\Cref{thm:oblivious} is larger than that obtained by~\Cref{thm:hardt-original}.
\end{proposition}
\noindent We prove this below.

\noindent\textbf{Base distribution}  Building on the intuition from the MNIST-CIFAR example, we assume a distribution $\basedist(x_1,x_2,y)$ over 
$\mathcal{X}_1\times\mathcal{X}_2 \times \mathcal{Y}$ where $\mathcal{X}_i$ is a set of size 10 and $\mathcal{Y} = \{y_{+},y_{-} \}$. In the base distribution $\basedist$ we assume that both $x_1,x_2$ are perfectly correlated with $y$ and so either is enough to determine the outcome. This can be seen as similar to MNIST-CIFAR where $x_{1}$ and $x_{2}$ correspond to the MNIST and CIFAR image respectively. We will also take the following:
\begin{enumerate}
    \item Both labels have equal probability, so $\basedist\left(y\right)=\frac{1}{2}$.
    \item Each set $\cX_i$ can be partitioned as: $\cX_{i}{=}\cX_{i+}\cup \cX_{i-}$ where $P(y{=}y_{+} \mid x_i {\in} \cX_{i+}) {=} 1$ and likewise for the other class. This is possible as each $x_i$ is perfectly correlated with $y$. We also take that $\lvert\cX\rvert_{i\star} = \lvert\cX\rvert_{i\star}$ for $i \in \{1,2 \}$ and $\star \in \{+,y_{-}\}$. 
    \item The collective controls 10\% of the data $\strength{=}0.1$, its target label is $y^{*}{=}y_{-}$, and the signal is $g\left(x_{1},x_{2}\right)=g\left(x_{1},C\right)$ where $C\in \cX_{2+}$ is constant.
    \item The learning algorithm $\mathcal{A}$ works, similarly to JTT, in two stages:
    \begin{enumerate}
        \item In the first stage, the algorithm learns a Bayes optimal classifier $f_{1}:\cX_{1}\rightarrow \cY$ that uses only $x_{1}$.
        \item In the second stage, the algorithm then stores all the $\left(x_{1},x_{2},y\right)$ triples that are misclassified by $f_{1}$ by saving the pair $\left(x_{1},x_{2}\right)$ in the error set $E$ and their label $y$ in a function $f_{E}:\cX_{1}\times \cX_{2}\rightarrow \cY$ that gets the pair $\left(x_{1},x_{2}\right)$ and outputs $y$. 
        \item The algorithm then outputs a classifier
        \[
        f\left(x_{1},x_{2}\right)=\begin{cases}
f_{E}\left(x_{1},x_{2}\right) & \left(x_{1},x_{2}\right)\in E\\
f_{1}\left(x_{1}\right) & \left(x_{1},x_{2}\right)\notin E
\end{cases}
        \]
    \end{enumerate}
\end{enumerate}

The second assumption defines the following 4 distributions.
\begin{enumerate}
    \item $P_{1+}\left(x_{1}\right)$ is a uniform discrete probability over the set $X_{1+}$ which contains 5 values that correspond to the label $y_{+}$, i.e. $P_{1+}\left(x_{1}\in X_{1+}\right)=\frac{1}{5}, P_{1}\left(x_{1}\notin X_{1+}\right)=0$.
    \item $P_{1-}\left(x_{1}\right)$ is a uniform discrete probability over the set $X_{1-}$ that correspond to the label $y_{-}$.
    \item $P_{2+}\left(x_{2}\right)$ is a uniform discrete probability over the set $X_{2+}$ that correspond to the label $y_{+}$.
    \item $P_{2-}\left(x_{2}\right)$ is a uniform discrete probability over the set $X_{2-}$ that correspond to the label $y_{-}$.
\end{enumerate}

Then we can get the full probabilities
\begin{align*}
\basedist\left(x_{1},x_{2}|y=y_{+}\right)&=P_{1+}\left(x_{1}\right)P_{2+}\left(x_{2}\right)\\\basedist\left(x_{1},x_{2}|y=y_{-}\right)&=P_{1-}\left(x_{1}\right)P_{2-}\left(x_{2}\right)\\\basedist\left(x_{1},x_{2}\right)&=\frac{1}{2}\left(P_{1+}\left(x_{1}\right)P_{2+}\left(x_{2}\right)+P_{1-}\left(x_{1}\right)P_{2-}\left(x_{2}\right)\right).    
\end{align*}

Let the collective signal be $g\left(x_{1},x_{2}\right)=g\left(x_{1},C\right)$ where $C$ is a constant and the target label is $y^{*}=y_{-}$.
Then the uniqueness of the signal is 
\begin{align*}
\xi&=\basedist\left(x_{1},C\right)=\sum_{x_{1}\in X_{1}}\left[P_{1+}\left(x_{1}\right)P_{2+}\left(C\right)P\left(y=y_{+}\right)+P_{1-}\left(x_{1}\right)P_{2-}\left(C\right)P\left(y=y_{-}\right)\right]\\&=\sum_{x_{1}\in X_{1}}\left[\frac{1}{2}\left(P_{1+}\left(x_{1}\right)P_{2+}\left(C\right)+P_{1-}\left(x_{1}\right)P_{2-}\left(C\right)\right)\right]\\&=\frac{1}{2}P_{2+}\left(C\right)\left(\sum_{x_{1}\in X_{1}}P_{1+}\left(x_{1}\right)\right)+\frac{1}{2}P_{2-}\left(C\right)\left(\sum_{x_{1}\in X_{1}}P_{1-}\left(x_{1}\right)\right)\\&=\frac{1}{2}\left(P_{2+}\left(C\right)+P_{2-}\left(C\right)\right)\\&=\frac{1}{2}\left(\frac{1}{5}+0\right)\\&=\frac{1}{10},
\end{align*}
and the sub-optimality gap is
\begin{align*}
\Delta&=\max_{x\in X^{*}}\max_{y}\left(\basedist\left(y|x\right)-\basedist\left(y^{*}|x\right)\right)\\&\geq\basedist\left(y_{+}|x_{1}\in X_{1+},x_{2}=C\right)-\basedist\left(y_{-}|x_{1}\in X_{1-},x_{2}=C\right)\\&=1.
 \end{align*}
Now, according to \cref{thm:hardt-original}, the lower bound of success when using Bayes optimal classifier with $\epsilon=0$ is
\[
S_{\text{Bayes}}\left(\alpha\right)\geq1-\frac{1-\alpha}{\alpha}\Delta\xi-\frac{\epsilon}{1-\epsilon}=1-\frac{9}{10}-0=0.1.
\]
\paragraph{Algorithmic bias}

The observed mixture distribution \mixdist{} now contains conflicting labels when $x_{1}\in\cX_{1+}$ and $x_{2}=C$, as it can be sampled from either the base distribution \basedist{} with $y{=}y_{+}$ or from the collective distribution \colldist{} with $y{=}y_{-}$.
As defined, the first-stage classifier $f_{1}$ of the learning algorithm $\mathcal{A}$ is Bayes optimal w.r.t $x_{1}$.
The Bayes optimal $f_{1}$, with the collective being small, will predict the $f_{1}\left(x_{1}\in\cX_{1+}\right)=y_{+}$ label as it is more probable to come from \basedist{}.
As a result, only the collective samples will be misclassified and will dominate the second stage classifier $f_{E}$ such that $f_{E}\left(x_{1},C\right)=y_{-}$.
Finally the algorithm will return the classifier $f=\mathcal{A}\left(\mixdist\right)$ where

\[
f\left(x_{1},x_{2}\right)=\begin{cases}
y^{*}=y_{-} & x_{2}=C\\
\arg\max_{y}\basedist\left(y|x_{1}\right) & \text{else}
\end{cases}.
\]
Compare this with a Bayes optimal classifier $f$ that uses both $x_{1}$ and $x_{2}$ equally when given a sample $\left(x_{1}\in X_{1+},x_{2}=C\right)$:
\begin{align*}
f\left(x_{1}\in X_{1+},x_{2}=C\right)&=\arg\max_{y}\mixdist\left(y|x_{1}\in X_{1+},x_{2}=C\right)\\&=\arg\max_{y}\begin{cases}
\mixdist\left(y=y_{+}|x_{1}\in X_{1+},x_{2}=C\right) & y=y_{+}\\
\mixdist\left(y=y_{-}|x_{1}\in X_{1+},x_{2}=C\right) & y=y_{-}
\end{cases}\\&=\arg\max_{y}\begin{cases}
\strength\colldist\left(y=y_{+}|x_{1}\in X_{1+},x_{2}=C\right)+\left(1-\strength\right)\basedist\left(y=y_{+}|x_{1}\in X_{1+},x_{2}=C\right)\\
\strength\colldist\left(y=y_{-}|x_{1}\in X_{1+},x_{2}=C\right)+\left(1-\strength\right)\basedist\left(y=y_{-}|x_{1}\in X_{1+},x_{2}=C\right)
\end{cases}\\&=\arg\max_{y}\begin{cases}
\strength\cdot0+\left(1-\strength\right)\cdot1 & y=y_{+}\\
\strength\cdot1+\left(1-\strength\right)\cdot0 & y=y_{-}
\end{cases}\\&=\arg\max_{y}\begin{cases}
0.9 & y=y_{+}\\
0.1 & y=y_{-}
\end{cases}=y_{+}
\end{align*}

\paragraph{Surrogate distribution and success bounds} 

Let \surrdist{} be the a distribution similar to \basedist{}, such that $\surrdist\left(y|x_{1}\right)=\basedist\left(y|x_{1}\right)$ but $x_{2}$ is discrete uniformly distributed over $N$ values, meaning $\surrdist\left(\mathcal{X^{*}}\right)=\surrdist\left(x_{2}=C\right)=\surrdist\left(x_{2}\right)=\frac{1}{N}$. Note than $N$ can be as large as we want, virtually resulting in $\surrdist\left(\mathcal{X^{*}}\right)\approx0$. In other words, large $N$ makes the signal almost $0$-unique.

Since $x_{2}$ in \surrdist{} is i.i.d. and is not correlated with $y$, the Bayes optimal classifier $f_{\surrmixed}$ on the mixture distribution \surrmixed{} will only will only $x_{1}$, or $x_{2}$ if it is equal to $C$:
\[
f_{\surrmixed}\left(x_{1},x_{2}\right)=\begin{cases}
y^{*} & x_{2}=C\\
\arg\max_{y}\surrdist\left(y|x_{1}\right) & \text{else}
\end{cases}.
\]
Then, from the definition of \surrdist{} it stems that $f_{\surrmixed}=h_{P_{\alpha}}$. Plugging that in the definition for $w$ (\cref{eq:obliv}) we get $w_{\surrdist}=0$. Now, the bound for success according to \cref{thm:oblivious}:
\[
S_{\text{bias}}\left(\alpha\right)\geq1-\frac{w}{1-w}-\frac{1-\alpha}{\alpha}Q_{0}\left(\mathcal{X^{*}}\right)=1-\frac{9}{N}\approx1 > S_{\text{Bayes}}
\]
Making the bound from \cref{thm:oblivious} tighter than the bound of \cref{thm:hardt-original}.

\begin{figure}[t]\centering
\begin{subfigure}[t]{0.48\linewidth}
    \centering
    \includegraphics[width=\linewidth]{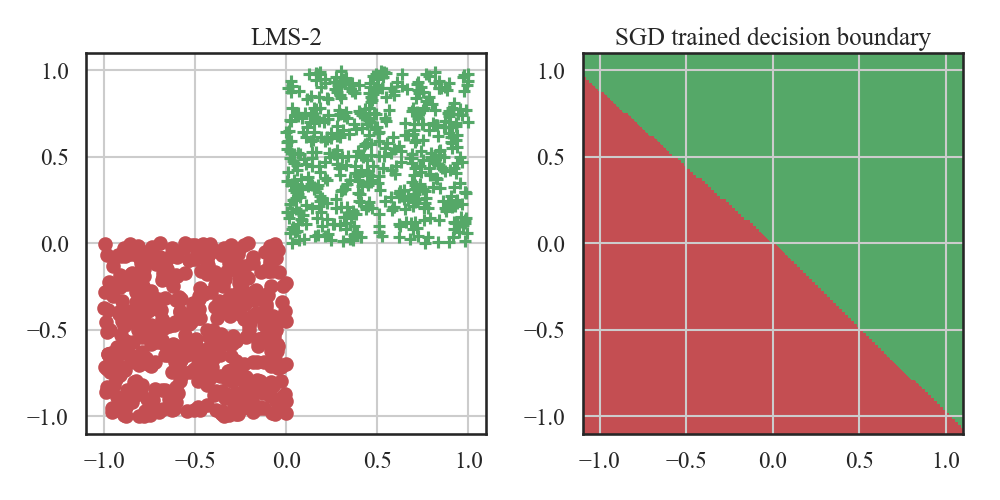}
    \caption{LMS-2}
    \label{fig:sb_lm2}
\end{subfigure}
\begin{subfigure}[t]{0.48\linewidth}
    \centering
    \includegraphics[width=\linewidth]{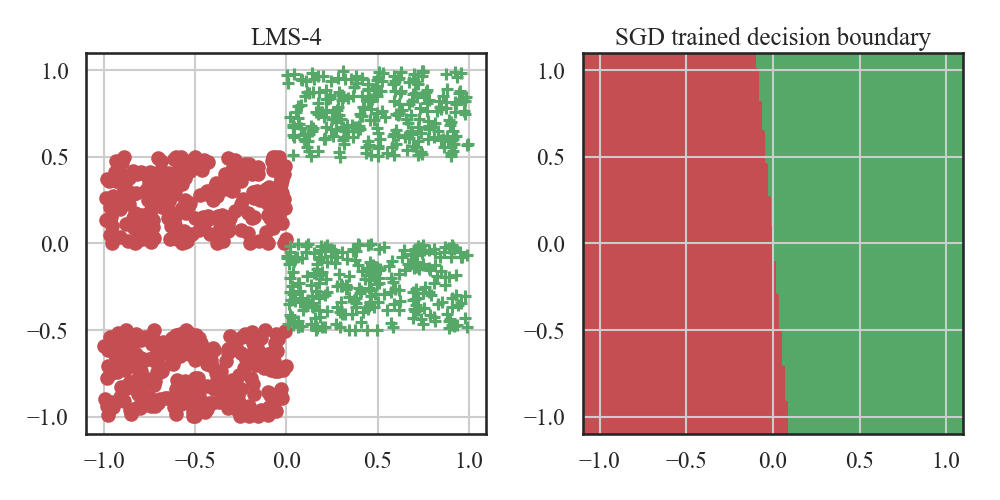}
    \caption{LMS-4}
    \label{fig:sb_lm4}
\end{subfigure}
\begin{subfigure}[t]{0.48\linewidth}
    \centering
    \includegraphics[width=\linewidth]{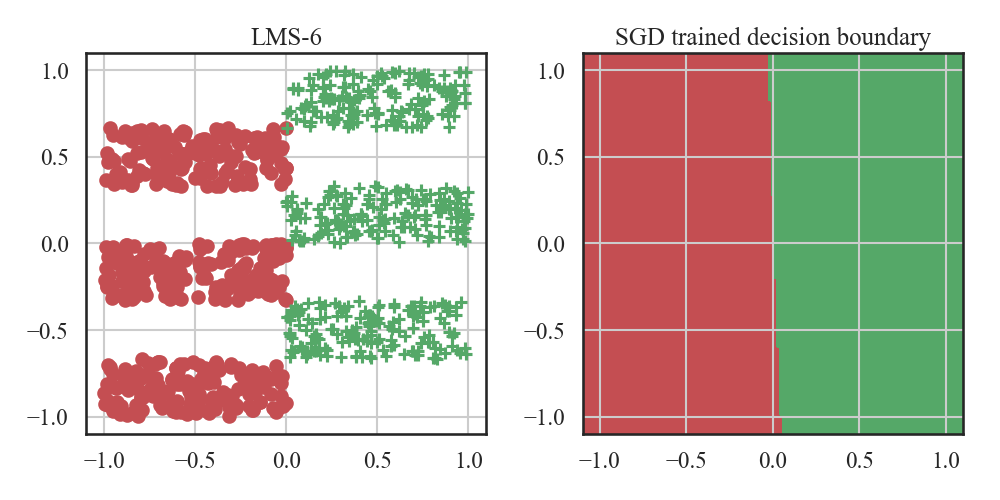}
    \caption{LMS-6}
    \label{fig:sb_lm6}
\end{subfigure}
\begin{subfigure}[t]{0.48\linewidth}
    \centering
    \includegraphics[width=\linewidth]{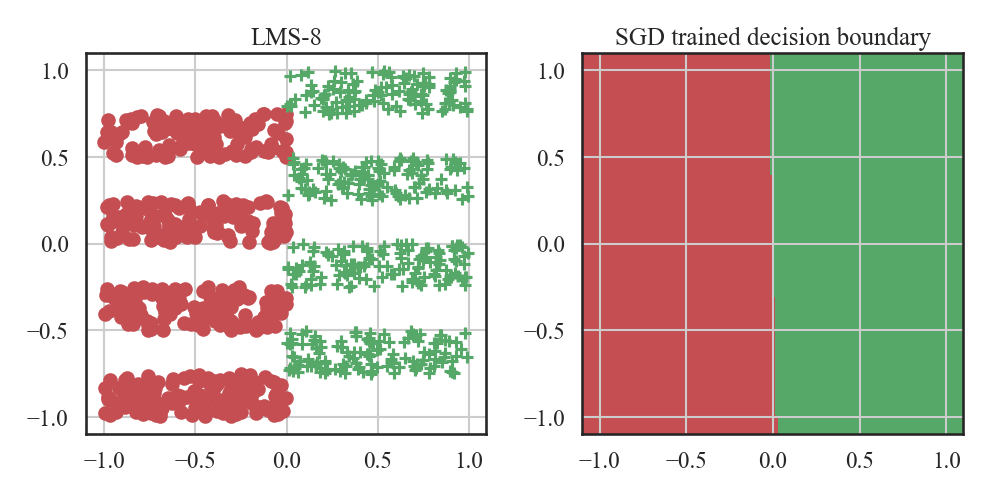}
    \caption{LMS-8}
    \label{fig:sb_lm8}
\end{subfigure}
\caption{Trained classifier on LMS-$k$ with no collective action. Left figure of each pair is a visualisation of the dataset, and the right figure of each pair is the decision boundary of the trained classifier.}%
\end{figure}

\section{Experiments}
\label{sec:experiment_details}
\noindent\textbf{Experimental details}
For the 2D datasets, we used an MLP with layers sizes of [64, 32, 16, 2] with ReLU activations.
For all image datasets we used the ResNet50 model.
In all experiments we used the PyTorch ADAM optimizer with the default parameters, a learning rate of $5\times10^{-4}$ and a batch size of $128$.
Each experiment was run multiple times with different random seeds, and in all figures the lines represent the means over the seeds, and the region around the lines is the $95\%$ confidence interval according to Student's $t$-distribution.

\noindent\textbf{Example of simplicity bias in action}
To show the effect of simplicity bias, here we repeat training a classifier on the LMS-$k$ dataset with no collective action for different $k$s.
\Cref{fig:sb_lm2,fig:sb_lm4,fig:sb_lm6,fig:sb_lm8} show how the decision boundary depends more on $x_1$ as $k$ grows.

\end{document}